\RequirePackage{booktabs}

\documentclass[pdflatex,sn-basic,Numbered]{sn-jnl}% Basic Springer Nature Reference Style/Chemistry Reference Style

\usepackage{soul}
\usepackage{adjustbox}
\usepackage{graphicx}%
\usepackage{multirow}%
\usepackage{amsmath,amssymb,amsfonts}%
\usepackage{amsthm}%
\usepackage{mathrsfs}%
\usepackage[title]{appendix}%
\usepackage{textcomp}%
\usepackage{manyfoot}%
\usepackage{booktabs}%
\usepackage{listings}%
\usepackage[ruled,vlined,linesnumbered]{algorithm2e}
\usepackage{algcompatible}
\usepackage{array}
\usepackage{xspace}
\usepackage{setspace}
\usepackage{hyperref}
\usepackage[dvipsnames]{xcolor}
\usepackage{tikz}
\usepackage{amsmath}
\usepackage{times,latexsym}
\usepackage{url}
\usepackage[T1]{fontenc}
\usepackage{pifont}

\usepackage[labelformat=simple,singlelinecheck=false]{subcaption}

\definecolor{darkgreen}{rgb}{0.0, 0.5, 0.0}

\newcommand{\xmark}{\ding{55}}

\makeatletter
\DeclareRobustCommand\onedot{\futurelet\@let@token\@onedot}
\def\@onedot{\ifx\@let@token.\else.\null\fi\xspace}

\begin{document}

\title[Article Title]{A Large-Scale Benchmark for Evaluating Large Language Models on Medical Question Answering in Romanian}


\author[1]{\fnm{Ana-Cristina} \sur{Rogoz{$^{\dagger,}$}}}
\author*[1]{\fnm{Radu Tudor} \sur{Ionescu{$^{\dagger,}$}}}\email{raducu.ionescu@gmail.com}
\author[2]{\fnm{Alexandra-Valentina}
\sur{Anghel}}
\author[2,3]{\fnm{Ionu\c{t}-Lucian} \sur{Antone-Iordache}}
\author[2,4]{\fnm{Simona} \sur{Coniac}}
\author[2,3,5]{\fnm{Andreea Iuliana} \sur{Ionescu}}

\affil[1]{\orgname{University of Bucharest}, \orgaddress{\street{14 Academiei}, \city{Bucharest}, \postcode{010014}, \country{Romania}}}

\affil[2]{\orgname{``Carol Davila'' University of Medicine and Pharmacy}, \orgaddress{\street{37 Dionisie Lupu}, \city{Bucharest}, \postcode{020021}, \country{Romania}}}

\affil[3]{\orgname{Col\c{t}ea Clinical Hospital}, \orgaddress{\street{1 Ion C. Br\u{a}tianu}, \city{Bucharest}, \postcode{030167}, \country{Romania}}}

\affil[4]{\orgname{Hospice Hope Bucharest}, \orgaddress{\street{121-123 T\u{a}m\^{a}ioarei}, \city{Bucharest}, \postcode{023642}, \country{Romania}}}

\affil[5]{\orgname{``Dan Furtun\u{a}'' Medical Center}, \orgaddress{\street{179 Splaiul Independen\c{t}ei}, \city{Bucharest}, \postcode{050099}, \country{Romania}}}

\affil[$\dagger$]{Equal contribution}


\abstract{
% Question answering (QA) is an actively studied topic, being a core natural language processing (NLP) task that needs to be addressed before achieving Artificial General Intelligence (AGI). However, the lack of QA datasets in specific domains and languages hinders the development of robust AI models able to generalize across various domains and languages. To this end, we
We introduce MedQARo, the first large-scale medical QA benchmark in Romanian, alongside a comprehensive evaluation of state-of-the-art large language models (LLMs). We construct a high-quality and large-scale dataset comprising 105,880 QA pairs about cancer patients from two medical centers. The questions regard medical case summaries of 1,242 patients, requiring both keyword extraction and reasoning. %MedQARo is the result of a time-consuming manual annotation process carried out by seven physicians specialized in oncology or radiotherapy, who spent a total of about 3,000 work hours to generate the QA pairs. 
Our benchmark contains both in-domain and cross-domain (cross-center and cross-cancer) test collections, enabling a precise assessment of generalization capabilities. We experiment with four open-source LLMs from distinct families of models on MedQARo. Each model is employed in two scenarios: zero-shot prompting and supervised fine-tuning. We also evaluate two state-of-the-art LLMs exposed only through APIs, namely GPT-5.2 and Gemini 3 Flash. Our results show that fine-tuned models significantly outperform zero-shot models, indicating that pretrained models fail to generalize on MedQARo. Our findings demonstrate the importance of both domain-specific and language-specific fine-tuning for reliable clinical QA in Romanian.
}

\keywords{Medical question answering, medical resource, Romanian language resource, medical LLMs.}



\maketitle

%\vspace{-0.1cm}
\section*{Introduction}
%\vspace{-0.1cm}
The development of robust question answering (QA) systems is an important goal in natural language processing (NLP), representing one of the prerequisites for building systems that reach human-level intelligence. Despite the significant advances in the area of Large Language Models (LLMs) \citep{OpenBioLLMs, touvron2023llama, jiang2023mistral7b, alpaca, kopf2024openassistant}, which are able to perform remarkably well across various QA tasks \citep{brown2020language, chowdhery2023palm, rajpurkar-etal-2016-squad, hendrycksmeasuring, openai2023gpt4, singhal2023large, singhal2023towards}, QA remains a challenging task within specialized domains, such as the medical field, or in low-resource languages, such as Romanian. For instance, the task of medical QA comes with unique challenges due to the complexity of clinical narratives and the necessity for deep understanding of medical terms. Moreover, there is a soaring demand for high accuracy, as mistakes can have serious consequences in daily medical practice assisted by AI agents. 

% \subsection*{Related Work}
% %\vspace{-0.1cm}
% % ✔️ TODO: - QA datasets in any language / multi-lingual
% \noindent\textbf{English and multi-lingual QA datasets.}

Question answering emerged as a fundamental task in NLP, with several datasets being developed to support research across various domains and languages. Popular QA datasets focusing on the English language, such as SQuAD \citep{rajpurkar-etal-2016-squad} and Natural Questions \citep{kwiatkowski-etal-2019-natural}, established the foundation for extractive and generative QA systems. These datasets showed the importance of large-scale high-quality annotations for training robust QA models. Later on, given the need for cross-language understanding, more and more datasets covering several languages started to arise. Some notable multilingual efforts include XQuAD \citep{artetxe-etal-2020-cross}, which extends SQuAD to 11 languages, and MLQA \citep{lewis-etal-2020-mlqa}, which provides QA datasets in seven languages with cross-lingual evaluation capabilities. Furthermore, TyDi QA \citep{clark-etal-2020-tydi} advanced multilingual QA by covering structurally different languages and focusing on information-seeking questions. However, these efforts did not include Romanian among the target languages.

\begin{table*}[t!]
\centering
\caption{Comparison between MedQARo and existing QA datasets across three dimensions, namely language coverage, domain focus, and dataset size. Existing QA datasets range from general-purpose English resources to multilingual datasets and domain-specific collections in areas such as legal, technical and medical fields. MedQARo is the first large-scale benchmark for medical QA in Romanian.}
\label{tab:dataset_comparison}
\resizebox{\textwidth}{!}{%
\setlength\tabcolsep{3pt}
\begin{tabular}{lllr}
\toprule
\textbf{Dataset} & \textbf{Language} & \textbf{Domain} & \textbf{\# QA pairs} \\
\midrule
SQuAD \citep{rajpurkar-etal-2016-squad} & English & General knowledge & 107,785 \\
% \hline
Natural Questions \citep{kwiatkowski-etal-2019-natural} & English & General knowledge & 307,373 \\
% \hline
XQuAD \citep{artetxe-etal-2020-cross} & Multilingual (11 languages) & \multirow{1}{*}{General knowledge} & \multirow{1}{*}{1,190} \\
% \hline
MLQA \citep{lewis-etal-2020-mlqa} & Multilingual (7 languages) & \multirow{1}{*}{General knowledge} & \multirow{1}{*}{17,000} \\
% \hline
TyDi QA \citep{clark-etal-2020-tydi} & Multilingual (11 languages)  & \multirow{1}{*}{Information-seeking} & \multirow{1}{*}{204,000} \\
\midrule
PubMedQA \citep{jin-etal-2019-pubmedqa} & English & Medical & 212,300 \\
% \hline
emrQA \citep{pampari-etal-2018-emrqa} & English & Clinical cases & 455,837 \\
% \hline
MedQA \citep{jin2020diseasedoespatienthave} & English & Medical exams & 12,723 \\
% \hline
MedMCQA \citep{pmlr-v174-pal22a} & English & Medical exams (India) & 193,100 \\
\midrule
JuRo \citep{craciun-etal-2025-graf} & Romanian & Legal & 10,836  \\
% \hline
RoITD \citep{nicolae2021roitd} & Romanian & Technical & 9,500 \\
% \hline
LiRo / XQuAD-Ro \citep{liro2021} & Romanian & General knowledge & 1,190 \\
RoMedQA \citep{dima-etal-2024-roqllama} & Romanian & Biology exams & 4,127 \\
\midrule
\textbf{MedQARo (ours)} & \textbf{Romanian} & \textbf{Clinical cases} & \textbf{105,880} \\
\bottomrule[0.8pt]%
\end{tabular}%
}
\end{table*}

% ✔️TODO: - Medical QA datasets
% \noindent\textbf{Medical QA datasets.}
Medical QA represents a challenging domain that requires deep understanding of a broad range of aspects, such as clinical terminology and medical reasoning. Several English medical QA datasets have been already developed to tackle the particularities of the medical domain. For instance, MedQA \citep{jin2020diseasedoespatienthave} provides over 12,000 multiple-choice questions from medical licensing exams across three countries, testing complex clinical reasoning abilities. Additionally, the emrQA dataset \citep{pampari-etal-2018-emrqa} introduces one of the largest QA medical resources with over 455,000 question-answer pairs automatically generated from electronic medical records. PubMedQA \citep{jin-etal-2019-pubmedqa} focuses on biomedical literature comprehension, with questions derived from research article titles and abstracts. MedMCQA \citep{pmlr-v174-pal22a} expands medical QA to include Indian medical entrance exams, with 193,100 questions across 21 medical subjects, demonstrating the importance of diverse medical knowledge sources. Despite its data source (Indian medical exams), the QA pairs contained by MedMCQA are in English.

While large-scale datasets such as PubMedQA \citep{jin-etal-2019-pubmedqa}, emrQA \citep{pampari-etal-2018-emrqa}, and MedQA \citep{jin2020diseasedoespatienthave} have driven significant progress in English medical QA, the availability of similar resources in other languages, especially underrepresented ones, remains limited. This hinders the development of LLMs that are robust to language and domain shift. % To the best of our knowledge, for the Romanian language, there are no prior efforts in the realm of large-scale QA resources for the medical domain.
% To the best of our knowledge, we are the first to introduce a large-scale QA dataset specifically tailored for the medical domain in Romanian. 
% In contrast to the significant advances in English medical QA, the landscape of non-English medical QA datasets remains scarce. 
To address this gap, we introduce MedQARo, the first large-scale benchmark for \textbf{Med}ical \textbf{Q}uestion \textbf{A}nswering in \textbf{Ro}manian. MedQARo contains original clinically-grounded content rather than translations, addressing a critical gap in current literature.

% % ✔️TODO: QA for Romanian
% \noindent\textbf{Romanian QA datasets.}
For the Romanian language, a few QA datasets have been independently developed. Two such datasets focus on domain specific understanding. One is the JuRo dataset \citep{craciun-etal-2025-graf}, which provides a collection of Romanian juridical texts with associated questions, and the other is the RoITD dataset \citep{nicolae2021roitd}, which addresses technical domain questions in the IT sector. Additionally, the LiRo benchmark \citep{liro2021}, which provides a comprehensive evaluation platform for Romanian NLP tasks, contributed with a subset of XQuAD \citep{artetxe-etal-2020-cross} for Romanian, consisting of translated versions of English questions. For advanced biology, \citet{dima-etal-2024-roqllama} published a small-scale dataset comprising 4,127 single-choice questions. Existing datasets for Romanian QA  are typically small (containing between 1,000 and 11,000 QA pairs). With 105,880 QA pairs, MedQARo is one order of magnitude larger than the largest dataset for Romanian QA.

% Multilingul (nr. limbi) 
% hline intre sectiuni 

% \noindent\textbf{MedQARo vs.~QA dataset landscape.}
In Table~\ref{tab:dataset_comparison}, we present a comprehensive comparison between MedQARo and existing QA datasets across different languages and domains. MedQARo fills in a gap in the landscape of medical QA resources by providing the first large-scale Romanian medical QA dataset with 105,880 question-answer pairs focused on oncology. While other Romanian QA datasets exist, none of them includes extractive and reasoning questions based on medical case summaries. This positions MedQARo as the first large-scale resource for advancing medical NLP capabilities for Romanian, an underrepresented language in the QA field.

% ✔️ Add: To the best of our knowledge, there is not medical QA dataset for Romanian.

% - Tabel similar cu Saroco: autor, limbi, domain (medical/other) / number of samples

\begin{figure}[t]
    \centering
    \includegraphics[width=0.5\linewidth]{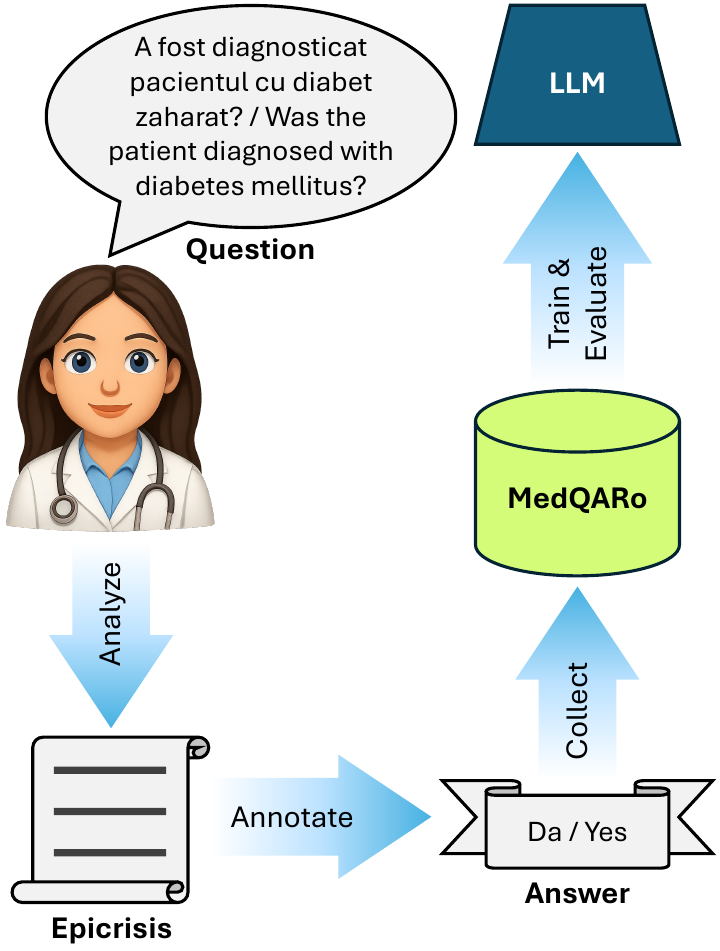}
    %\vspace{-0.2cm}
    \caption{An image illustrating our dataset creation and model benchmarking stages. In the annotation stage, physicians analyze epicrises in order to find the correct answer for a given question. Upon collecting question and answer QA pairs, we benchmark zero-shot and fine-tuned LLMs on the medical question answering task, using two evaluation scenarios, namely in-domain and cross-domain.}
    \label{fig_teaser}
    %\vspace{-0.2cm}
\end{figure}

As illustrated in Figure \ref{fig_teaser}, our benchmark creation process involves two phases, (i) manual data collection and annotation, and (ii) model benchmarking. The dataset comprises 105,880 high-quality QA pairs from real-world clinical records of 1,242 oncology patients (796 patients with breast cancer, 215 patients with lung cancer, and 231 patients with other cancers). The QA pairs are the results of a manual annotation process carried out by physicians specialized in oncology and radiotherapy, who collectively spent nearly 3,000 hours to generate the pairs, ensuring both linguistic quality and medical correctness. MedQARo includes 76,416 QA pairs about breast cancer patients, 26,230 about lung cancer patients and 3,234 about patients with other cancer types, with questions grounded in medical case summaries (epicrises).

% To ensure the high quality of our dataset, all patient-related clinical data was manually authored by a team of seven physicians, ensuring both linguistic quality and medical correctness.

%there are prior efforts in the realm of QA such as the work done by \citet{craciun-etal-2025-graf}, \citet{liro2021} and \citet{nicolae2021roitd}. However, none of them tackle the problem of QA for the medical domain. 

To prevent data leakage across splits and support fair model development and evaluation, we provide official training, validation, and test splits at the patient level, i.e.~a medical record can appear in only one of the splits. Moreover, we provide both in-domain and cross-domain test collections, where the cross-domain test set comprises patients from a distinct medical center, who suffer from one of five cancer types (head and neck, melanoma, renal cell carcinoma, bladder cancer, hepatocellular carcinoma) that are not available during training. Hence, MedQARo enables a precise assessment of generalization capabilities across different medical centers and cancer types. Oncologists, especially residents, could greatly benefit from using LLMs to extract and reason about the information included in epicrises, when they need to recommend diagnostic tools or treatment lines for their patients. MedQARo provides the necessary means to develop such LLMs, that could be deployed in daily medical practice. We benchmark four models on MedQARo, namely two LLMs specialized on Romanian (RoLLaMA2-7B and RoMistral-7B) \citep{masala2024vorbecstiromanecsterecipetrain}, a long-context LLM (Phi-4-mini-instruct) \citep{phi4mini}, and an LLM tuned on biomedical data (LLaMA3-OpenBioLLM-8B) \citep{OpenBioLLMs}. Each model is assessed in both zero-shot and supervised fine-tuning setups, while testing various prompt formats. In addition, we assess two state-of-the-art LLMs in the zero-shot setting, which are only accessible through paid APIs, namely GPT-5.2 and Gemini 3 Flash. Our results show that all LLMs significantly benefit from fine-tuning, whereas zero-shot variants perform poorly. These findings emphasize the critical importance of both domain and language adaptation for reliable clinical QA in Romanian. Among all evaluated models, the fine-tuned RoMistral-7B achieves the best performance, outperforming all the other approaches. However, the best-performing model achieves an F1 score of only $0.671$ on the in-domain test set, suggesting that MedQARo is a challenging dataset, highlighting the need for the development of more robust models, capable of better generalizing to specialized domains and low-resource languages.

In summary, our contributions are fivefold:
\begin{itemize}
    \item We introduce \textbf{MedQARo}, the first large-scale medical QA dataset in Romanian, comprising 105,880 QA instances obtained with the help of trained physicians.
    % \item We construct the dataset using real clinical case summaries from 1,011 patients diagnosed with either breast or lung cancer, paired with 48 and 61 expert-crafted questions, respectively.
    %\item We provide an official dataset split at the patient level, yielding 71,772 training, 15,328 validation, and 15,546 test samples, supporting both model training and robust evaluation.
    \item We benchmark four LLMs from distinct families of models across multiple configurations and report comprehensive performance metrics under zero-shot and fine-tuning configurations, in both in-domain and cross-domain settings.
    \item We demonstrate that domain-specific and language-specific adaptation is crucial for reliable clinical QA in low-resource languages such as Romanian.
    \item We show that state-of-the-art API-based LLMs, such as GPT-5.2 and Gemini 3 Flash, perform far worse than smaller fine-tuned LLMs, further confirming the importance of fine-tuning.
    \item We release the dataset and code to reproduce the results at \url{https://github.com/ana-rogoz/MedQARo}, where patient sensitive data is fully anonymized. 
\end{itemize}

\section*{Results}
%\vspace{-0.1cm}

We conduct experiments with the chosen methods on the MedQARo dataset, using our official data split. We start by tuning various hyperparameters involved in the fine-tuning process, aiming to find the optimal hyperparameter setup for all LLMs. We next evaluate two different prompt formats, to determine which prompt format leads to higher performance. For the best performing prompt structure, we conduct several experiments across multiple LLMs, using various prompt lengths. The objectives of our experiments are: (i) to capture how Romanian-adapted models compare with general-purpose and domain-specific alternatives, (ii) to determine how fine-tuning performs against zero-shot inference, (iii) to establish how fine-tuned or zero-shot LLMs compare against trivial baselines, (iv) to investigate how the context length influences model performance, particularly for models with extended context capabilities, and (v) to determine the generalization capacity of LLMs across distinct data distributions, where patients come from different medical centers, and suffer from different cancer types. 

%\vspace{-0.1cm}
\subsection*{Evaluation}
%\vspace{-0.1cm}

We employ four alternative evaluation metrics: 
\begin{itemize}
    \item \textbf{F1 score:} The first metric that we take into consideration is the F1 score, which measures precision and recall at the token level. It provides a balanced view of how well models identify certain parts of the answer.
    \item \textbf{Exact match:} We also consider the exact match (EM) score, which represents the percentage of predictions that perfectly match the ground-truth answer. Although this metric is the most demanding, it is crucial in medical contexts where small wording differences can change meaning.
    \item \textbf{BLEU:} Because MedQARo contains answers with more than one token, we consider the BLEU score \citep{papineni2002bleu}, which measures n-gram overlap to assess fluency and lexical similarity. When the predicted answer includes only a subset of the tokens from the ground-truth answer, BLEU can provide a representative measure by considering partial n-gram matches.
    \item \textbf{METEOR:} Lastly, we include METEOR \citep{banerjee2005meteor}, which offers a more flexible evaluation by accounting for synonyms, stemming, and word order variations. These aspects are particularly valuable given the rich morphology and flexible syntax of Romanian.
\end{itemize}

%\vspace{-0.1cm}
\subsection*{Model Training Setup and Hyperparameter Tuning}
%\vspace{-0.1cm}

% \noindent\textbf{LoRA adaptation details.}
All models are fine-tuned using a LoRA strategy applied to the causal language modeling objective. For RoLLaMA2-7B-Instruct, RoMistral-7B-Instruct, and LLaMA3-OpenBioLLM-8B, LoRA adapters are injected into the self-attention projection layers (query, key, value, and output projections), following the default configuration for these architectures.
For Phi-4-mini-instruct, due to architectural differences, LoRA is applied to the fused QKV projection layer and the output projection layer, respectively.
In all cases, the original model parameters are kept frozen, and only the LoRA adapters are trainable.
Feed-forward layers, token embeddings, layer normalization parameters, language modeling heads, and all bias terms remain frozen throughout training.
This attention-only adaptation results in updating approximately 0.04-0.10\% of the total number of weights, depending on the core architecture.

% \noindent\textbf{Hyperparameter tuning.} 

We employ consistent hyperparameter settings across all models to ensure a fair comparison among LLM families. All models are fine-tuned using the AdamW \citep{loshchilov2019decoupled} optimizer. Due to memory constraints, we set a per-device batch size of one sample with gradient accumulation over $8$ steps, resulting in an effective mini-batch size of $8$ samples. Training is carried out for a maximum of $2$ epochs, using Brain Floating Point (BFLOAT16) precision to optimize memory usage, while maintaining numerical stability. 
The learning rate, the dropout rate and the LoRA configuration are established via grid search on the validation set. The range for the initial learning rate is $10^{-3}$ to $10^{-6}$. We employ a cosine learning rate scheduler with $100$ warm-up steps. Dropout is applied with drop rates between $0$ and $0.2$, using a step of $0.05$. For the LoRA rank $r$, we look for an optimal value in the set $\{4, 6, 8, 10\}$. For the LoRA scaling factor, we consider values in the set $\{8, 16, 32\}$. The optimal configuration is based on a learning rate of $2\cdot 10^{-5}$, a dropout rate of $0.05$, a LoRA rank $r$ of $8$, and a LoRA scaling factor $\alpha$ of $16$.

\begin{table*}[t!]
\centering
\caption{Preliminary results with alternative prompt formats for RoLLaMA based on 2,048 tokens. We compare two prompt formats: epicrisis + question + answer (\texttt{E+Q+A}) vs.~question + epicrisis + answer (\texttt{Q+E+A}). The \texttt{Q+E+A} format yields higher performance across all evaluation metrics. The best results are highlighted in \textcolor{darkgreen}{\textbf{bold green}}.}
\label{tab:preliminary}
\begin{adjustbox}{width=\textwidth}
\setlength\tabcolsep{2.9pt}
\begin{tabular}{lcccccccccc}
\toprule
\multirow{2.5}{*}{\textbf{Model}} & \multirow{2.5}{*}{\textbf{Prompt format}} & \multicolumn{4}{c}{\textbf{Validation}} & & \multicolumn{4}{c}{\textbf{In-domain test}} \\
\cmidrule{3-6}
\cmidrule{8-11}
&  & \textbf{F1} & \textbf{EM} & \textbf{BLEU} & \textbf{METEOR} & & \textbf{F1} & \textbf{EM} & \textbf{BLEU} & \textbf{METEOR} \\
\midrule
\multirow{2}{*}{\textbf{RoLLaMA2-7B}} 
 & \texttt{E+Q+A} & 0.4012	& 0.3082	& 0.2820 &	0.2843	& & 0.4087	& 0.319	& 0.2915	& 0.2975 \\
 & \texttt{Q+E+A} & \textcolor{darkgreen}{\textbf{0.5674}} &	\textcolor{darkgreen}{\textbf{0.5572}} &	\textcolor{darkgreen}{\textbf{0.5056}}	& \textcolor{darkgreen}{\textbf{0.3645}} & &	\textcolor{darkgreen}{\textbf{0.5759}} & 	\textcolor{darkgreen}{\textbf{0.5708}}	& \textcolor{darkgreen}{\textbf{0.5211}}	& \textcolor{darkgreen}{\textbf{0.3688}} \\
\bottomrule[0.8pt]
\end{tabular}
\end{adjustbox}
\end{table*}

\subsection*{Preliminary Results on Prompt Formatting}

We conduct a preliminary experiment to assess the impact of the prompt structure on QA performance. Specifically, we compare two input formulations for the RoLLaMA2-7B-Instruct model: epicrisis + question + answer (\texttt{E+Q+A}) vs.~question + epicrisis + answer (\texttt{Q+E+A}). Both configurations are evaluated with 4,096 input tokens. As shown in Table~\ref{tab:preliminary}, the \texttt{Q+E+A} format consistently outperforms the \texttt{E+Q+A} format across all evaluation metrics. As confirmed by other studies \cite{Barbero-Arxiv-2025}, the first tokens tend to receive more attention, which might explain why placing the question at the beginning helps the model focus more on the question that needs to be answered. Given these results, we decided to run all subsequent experiments using the \texttt{Q+E+A} prompt structure.

% majority answer 
% fine tuned vs. non-fine tuned 

\begin{table*}[t!]
\centering
\caption{In-domain results of multiple open-source LLMs based on various configurations vs.~two trivial baselines, on MedQARo. For each LLM, there is a zero-shot version and several fine-tuned versions, with various prompt lengths. F1, Exact Match (EM), BLEU, and METEOR scores are reported on both validation and test sets. Fine-tuned models consistently outperform zero-shot versions. The best results for each LLM are highlighted in \textcolor{darkgreen}{\textbf{bold green}}.}
\label{tab:results}
\begin{adjustbox}{width=\textwidth}
\setlength\tabcolsep{3.5pt}
\begin{tabular}{lcc cccc c cccc}
\toprule
\multirow{2.5}{*}{\textbf{Model}} & \multirow{2.5}{*}{\textbf{Fine-tuned}} & \multirow{2.5}{*}{\textbf{\#Tokens}} & \multicolumn{4}{c}{\textbf{Validation}} & & \multicolumn{4}{c}{\textbf{In-domain test}} \\
\cmidrule{4-7}
\cmidrule{9-12}
& & & \textbf{F1} & \textbf{EM} & \textbf{BLEU} & \textbf{METEOR} & & \textbf{F1} & \textbf{EM} & \textbf{BLEU} & \textbf{METEOR} \\
\midrule
\multirow{3}{*}{\textbf{Random token selector}} 
& \multirow{3}{*}{N/A}  & 1,024 & 0.0023 & 0.0006 & 0.0006 & 0.0013 & & 0.0019 & 0.0001 & 0.0004 & 0.0011 \\
&  & 2,048 & 0.0021 & 0.0004 & 0.0004 & 0.0013 & & 0.0024 & 0.0003 & 0.0005 & 0.0015 \\
&  & 4,096 & 0.0022 & 0.0005 & 0.0006 & 0.0014 & & 0.0026 & 0.0003 & 0.0005 & 0.0015 \\
\midrule
{\textbf{Majority answer}} 
& {N/A}  & N/A & 0.2091 & 0.1786 & 0.1831 & 0.1153 & & 0.2148 & 0.1838 & 0.1884 & 0.1183 \\

\midrule
\multirow{4.5}{*}{\textbf{RoLLaMA2-7B}} 
& \textcolor{red}{\xmark} & 2,048 & 0.0188 & 0.0015 & 0.0024 & 0.0146 & & 0.0230 & 0.0025 & 0.0023 & 0.0166 \\
\cmidrule{2-12}
& \textcolor{darkgreen}{\checkmark} & 1,024 & 0.5289	& 0.5529	& 0.4991	& 0.3312	& &  0.5316	& 0.5551	& 0.4990	& 0.3305 \\
& \textcolor{darkgreen}{\checkmark} & 2,048 & \textcolor{darkgreen}{\textbf{0.5674}} &	\textcolor{darkgreen}{\textbf{0.5572}} &	\textcolor{darkgreen}{\textbf{0.5056}}	& \textcolor{darkgreen}{\textbf{0.3645}} & &	\textcolor{darkgreen}{\textbf{0.5759}} & 	\textcolor{darkgreen}{\textbf{0.5708}}	& \textcolor{darkgreen}{\textbf{0.5211}}	& \textcolor{darkgreen}{\textbf{0.3688}} \\
& \textcolor{darkgreen}{\checkmark} & 4,096 & 0.3290	& 0.3055	& 0.2786	& 0.1980	& & 0.3267	& 0.3267	& 0.3267	& 0.1980 \\

\midrule
\multirow{3.5}{*}{\textbf{RoMistral-7B}} 
& \textcolor{red}{\xmark} & 4,096 & 0.0715 & 0.0430 & 0.0234 & 0.0438 & & 0.0605 & 0.0350 & 0.0229 & 0.0403 \\
\cmidrule{2-12}
& \textcolor{darkgreen}{\checkmark} & 2,048 & \textcolor{darkgreen}{\textbf{0.6731}} &	\textcolor{darkgreen}{\textbf{0.6977}}	& \textcolor{darkgreen}{\textbf{0.6458}}	& \textcolor{darkgreen}{\textbf{0.4170}} & & 	\textcolor{darkgreen}{\textbf{0.6716}}	& \textcolor{darkgreen}{\textbf{0.6956}}	& \textcolor{darkgreen}{\textbf{0.6445}}	& \textcolor{darkgreen}{\textbf{0.4180}} \\
& \textcolor{darkgreen}{\checkmark} & 4,096 & 0.6568	& 0.6882	& 0.6405	& 0.4078	& & 0.6587	& 0.6864	& 0.6391	& 0.4068 \\

\midrule
\multirow{7.5}{*}{\textbf{Phi-4-mini}} 
& \textcolor{red}{\xmark} & 3,072 & 0.0078 & 0.0000 & 0.0007 & 0.0075 & & 0.0048 & 0.0002 & 0.0005 & 0.0043 \\
\cline{2-12}
& \textcolor{darkgreen}{\checkmark} & 2,048 & \textcolor{darkgreen}{\textbf{0.6267}} &	\textcolor{darkgreen}{\textbf{0.6575}}	& \textcolor{darkgreen}{\textbf{0.5977}}	& \textcolor{darkgreen}{\textbf{0.3897}} &	& \textcolor{darkgreen}{\textbf{0.6291}}	& \textcolor{darkgreen}{\textbf{0.6593}} &	\textcolor{darkgreen}{\textbf{0.5999}} &	\textcolor{darkgreen}{\textbf{0.3916}} \\
& \textcolor{darkgreen}{\checkmark} & 3,072 & 0.6193 &	0.6493	& 0.5884	& 0.3853 &	& 0.6251 &	0.6529	& 0.5941 &	0.3876 \\
& \textcolor{darkgreen}{\checkmark} & 4,096 & 0.4806	& 0.4905	& 0.4463	& 0.2886	& & 0.4809 & 0.4897	& 0.4477 &	0.2893 \\
& \textcolor{darkgreen}{\checkmark} & 8,192 & 0.5234	& 0.5367	& 0.4912	& 0.3127	& & 0.5263	& 0.5374	& 0.4913	& 0.3156 \\
& \textcolor{darkgreen}{\checkmark} & 16,384 & 0.5275	& 0.5430	& 0.4963	& 0.3165 & &	0.5297	& 0.5415	& 0.4945	& 0.3183 \\
& \textcolor{darkgreen}{\checkmark} & 32,768 & 0.5277	& 0.5434	& 0.4970	& 0.3166	& & 0.5299	& 0.5418	& 0.4948 &	0.3186 \\

\midrule
\multirow{3.5}{*}{\textbf{OpenBioLLM-8B}} 
& \textcolor{red}{\xmark} & 2,048 & 0.0140 & 0.0004 & 0.0014 & 0.0156 & & 0.0119 & 0.0004 & 0.0015 & 0.0132 \\
\cmidrule{2-12}
& \textcolor{darkgreen}{\checkmark} & 2,048 & \textcolor{darkgreen}{\textbf{0.5465}} &	\textcolor{darkgreen}{\textbf{0.5730}}	& \textcolor{darkgreen}{\textbf{0.5239}} &	\textcolor{darkgreen}{\textbf{0.3389}} & &	\textcolor{darkgreen}{\textbf{0.5520}} &	\textcolor{darkgreen}{\textbf{0.5785}}	& \textcolor{darkgreen}{\textbf{0.5287}}	& \textcolor{darkgreen}{\textbf{0.3432}} \\
& \textcolor{darkgreen}{\checkmark} & 4,096 & 0.5217 &	0.5322 &	0.4873 &	0.3180 & &	0.5208	& 0.5338	& 0.4882 &	0.3203 \\
\bottomrule[0.8pt]
\end{tabular}
\end{adjustbox}
\end{table*}

\begin{table*}[t!]
\centering
\caption{Cross-domain (cross-center and cross-diagnosis) results of multiple open-source LLMs based on various configurations vs.~the majority answer baseline, on MedQARo. For each LLM, there is a zero-shot version and several fine-tuned versions, with various prompt lengths. Models are evaluated in terms of F1, EM, BLEU, and METEOR on a single held-out set comprising 231 patients and 3,234 QA pairs. The best results for each LLM are highlighted in \textcolor{darkgreen}{\textbf{bold green}}.}
\label{tab:secondary_results}
%\begin{adjustbox}{width=\textwidth}
\setlength\tabcolsep{3.5pt}
\begin{tabular}{lcc cccc}
\toprule
\multirow{2.5}{*}{\textbf{Model}} & \multirow{2.5}{*}{\textbf{Fine-tuned}} & \multirow{2.5}{*}{\textbf{\#Tokens}} & \multicolumn{4}{c}{\textbf{Cross-domain test}} \\
\cmidrule(lr){4-7}
& & & \textbf{F1} & \textbf{EM} & \textbf{BLEU} & \textbf{METEOR} \\
\midrule

\textbf{Majority answer} & N/A & N/A & 0.0909 & 0.0544 & 0.0544 & 0.0402 \\
\midrule

% ---------------- RoLLaMA2-7B ----------------
\multirow{4.5}{*}{\textbf{RoLLaMA2-7B}}
& \textcolor{red}{\xmark} & 2,048 & 0.0746	& 0.0087	& 0.0138	& 0.0617 \\
\cmidrule{2-7}
& \textcolor{darkgreen}{\checkmark} & 1,024 & 0.2919 &	0.2313 &	0.2030	& 0.1425 \\
& \textcolor{darkgreen}{\checkmark} & 2,048 & \textcolor{darkgreen}{\textbf{0.3036}} &	\textcolor{darkgreen}{\textbf{0.2319}}	& \textcolor{darkgreen}{\textbf{0.2126}}	& \textcolor{darkgreen}{\textbf{0.1526}} \\
& \textcolor{darkgreen}{\checkmark} & 4,096 & 0.1205 & 	0.0816	& 0.0732	& 0.0614 \\
\midrule

% ---------------- RoMistral ----------------
\multirow{3.5}{*}{\textbf{RoMistral-7B}}
& \textcolor{red}{\xmark} & 2,048 & 0.1286	& 0.0544	& 0.0469	& 0.0959 \\
\cmidrule{2-7}
& \textcolor{darkgreen}{\checkmark} & 2,048 & 0.4387 &	0.3605	& 0.3688	& 0.2225 \\
& \textcolor{darkgreen}{\checkmark} & 4,096 & \textcolor{darkgreen}{\textbf{0.4454}} &	\textcolor{darkgreen}{\textbf{0.3624}}	& \textcolor{darkgreen}{\textbf{0.3696}}	& \textcolor{darkgreen}{\textbf{0.2248}} \\
\midrule

% ---------------- Phi-4-mini ----------------
\multirow{7.5}{*}{\textbf{Phi-4-mini}}
& \textcolor{red}{\xmark} & 2,048 & 0.0120	& 0.000 &	0.0020	& 0.0112 \\
\cmidrule{2-7}
& \textcolor{darkgreen}{\checkmark} & 2,048 & \textcolor{darkgreen}{\textbf{0.3648}}	& \textcolor{darkgreen}{\textbf{0.3092}}	& \textcolor{darkgreen}{\textbf{0.3120}}	& \textcolor{darkgreen}{\textbf{0.1865}} \\
& \textcolor{darkgreen}{\checkmark} & 3,072 & 0.3446	& 0.2993	& 0.3038 &	0.1762 \\
& \textcolor{darkgreen}{\checkmark} & 4,096 & 0.3052	& 0.2669	& 0.2670	& 0.1587 \\
& \textcolor{darkgreen}{\checkmark} & 8,192 & 0.3663	& 0.3101	& 0.3061	& 0.1883 \\
& \textcolor{darkgreen}{\checkmark} & 16,384 & 0.3621	& 0.3061	& 0.3004	& 0.1843 \\
& \textcolor{darkgreen}{\checkmark} & 32,768 & 0.3613	& 0.3061	& 0.3018	& 0.1838 \\
\midrule

% ---------------- OpenBioLLM ----------------
\multirow{3.5}{*}{\textbf{OpenBioLLM-8B}}
& \textcolor{red}{\xmark} & 2,048 & 0.0196	& 0.0006	& 0.0029	& 0.0183 \\
\cmidrule{2-7}
& \textcolor{darkgreen}{\checkmark} & 2,048 & \textcolor{darkgreen}{\textbf{0.3165}}	& \textcolor{darkgreen}{\textbf{0.2638}}	& \textcolor{darkgreen}{\textbf{0.2682}} &	\textcolor{darkgreen}{\textbf{0.1606}} \\
& \textcolor{darkgreen}{\checkmark} & 4,096 & 0.2932	& 0.2440 &	0.2425	& 0.1503 \\
\bottomrule[0.8pt]
\end{tabular}
%\end{adjustbox}
\end{table*}

%\vspace{-0.1cm}
\subsection*{Main Results}
%\vspace{-0.1cm}

In Table~\ref{tab:results}, we present the results obtained by the chosen open-source LLMs vs.~two baselines on MedQARo. We evaluate all four LLMs under both zero-shot prompting and supervised fine-tuning setups, reporting performance metrics on both validation and in-domain test sets. 

We first compare baselines vs.~LLMs. In the zero-shot setup, the performance levels reached by the various LLMs indicate that out-of-the-box models can surpass the naive random token selector. However, the majority answer baseline, which leverages knowledge about the answer distribution in training data, significantly outperforms all zero-shot variants. This is a first hint that zero-shot models are not able to generalize to the medical QA task in Romanian, regardless of their prior language or domain adaptation. In contrast, each fine-tuned LLM version surpasses the two baseline models, irrespective of prompt length and prior model adaption. This observation holds for all LLM architectures, indicating that the fine-tuning process is generally useful.

We next compare zero-shot vs.~fine-tuned LLMs. In general, zero-shot models exhibit significantly lower performance than their fine-tuned counterparts. Furthermore, for various performance metrics and LLM architectures, we observe that the scores can increase by an order of magnitude when going from zero-shot prompting to fine-tuning. The comparison between zero-shot and fine-tuning setups highlights the critical importance of task-specific fine-tuning for clinical QA, confirming that prior language-specific or domain-specific adaptation is insufficient to accurately address the task.

\begin{table*}[t!]
\centering
\caption{Comparison of best-performing model variants against state-of-the-art baselines on a randomly sampled subset of 100 QA pairs from the in-domain test set. State-of-the-art API-based LLMs are evaluated in the zero-shot setting, while the remaining models are either zero-shot prompted or fine-tuned on MedQARo. The best F1, EM, BLEU, and METEOR scores are highlighted in \textcolor{darkgreen}{\textbf{bold green}}.}
\label{tab:testset_random100_sota}
%\begin{adjustbox}{width=\textwidth}
%\setlength\tabcolsep{10pt}
%\renewcommand{\arraystretch}{1.3}
\begin{tabular}{l c cccc}
\toprule
\textbf{Model} & \textbf{Fine-tuned} & \textbf{F1} & \textbf{EM} & \textbf{BLEU} & \textbf{METEOR} \\
\midrule

\textbf{GPT-5.2}         
& \textcolor{red}{\xmark} 
& 0.2415 & 0.1900 & 0.1339 & 0.1169 \\

\textbf{Gemini 3 Flash} 
& \textcolor{red}{\xmark} 
& 0.2041 & 0.1400 & 0.0099 & 0.1039 \\
\midrule

% ---------------- RoLLaMA2-7B ----------------
\multirow{2}{*}{\textbf{RoLLaMA2-7B}}
& \textcolor{red}{\xmark} 
& 0.0257  & 0.0000 & 0.0009 & 0.0187 \\
& \textcolor{darkgreen}{\checkmark} 
& 0.5213	& 0.5100	& 0.4730 &	0.3259 \\
\midrule
% ---------------- RoMistral-7B ----------------
\multirow{2}{*}{\textbf{RoMistral-7B}}
& \textcolor{red}{\xmark} 
& 0.0892 & 0.0700 & 0.0015 & 0.0568 \\
& \textcolor{darkgreen}{\checkmark} 
& \textcolor{darkgreen}{\textbf{0.6309}} &	\textcolor{darkgreen}{\textbf{0.6700}} &	\textcolor{darkgreen}{\textbf{0.6402}} & 	\textcolor{darkgreen}{\textbf{0.3862}} \\

\midrule

% ---------------- Phi-4-mini ----------------
\multirow{2}{*}{\textbf{Phi-4-mini}}
& \textcolor{red}{\xmark} 
& 0.0012 & 0.0000 & 0.0007 & 0.0045 \\
& \textcolor{darkgreen}{\checkmark} 
& 0.5867	& 0.5900	& 0.5413	& 0.3482 \\
\midrule

% ---------------- OpenBioLLM-8B ----------------
\multirow{2}{*}{\textbf{OpenBioLLM-8B}}
& \textcolor{red}{\xmark} 
& 0.0025 & 0.0000 & 0.0003 & 0.0031 \\
& \textcolor{darkgreen}{\checkmark} 
& 0.5180 &	0.5400 &	0.4997	& 0.3198 \\
\bottomrule[0.8pt]
\end{tabular}
%\end{adjustbox}
\end{table*}

We now analyze the cross-domain evaluation results. In Table~\ref{tab:secondary_results}, we report the results obtained on the cross-domain test set, which includes patients from a different medical center, who suffer from other cancer types than those observed during training, thus representing a substantially harder scenario for fine-tuned models. As expected, all fine-tuned models exhibit a noticeable performance drop with respect to the in-domain test set (see Table~\ref{tab:results}), confirming that cross-center and cross-diagnosis evaluation poses significant challenges. The majority answer baseline achieves low but non-trivial scores, highlighting that naive priors still provide a competitive lower bound in this setting. Zero-shot variants remain largely ineffective, in some cases performing close to the majority answer baseline, which further emphasizes that neither prior Romanian language specialization nor biomedical pretraining alone is sufficient to handle distribution shifts.

RoMistral-7B demonstrates the strongest robustness among all fine-tuned LLMs, achieving the highest F1 and EM scores with both prompt lengths. Phi-4-mini-instruct, remains the second best performer in both in-domain and cross-domain settings, while experiencing pronounced degradation when going from the in-domain setup to the cross-domain setup. OpenBioLLM-8B benefits from fine-tuning, but fails to outperform the Romanian-adapted RoMistral-7B model or the generic Phi-4-mini-instruct model, suggesting that English biomedical pretraining does not adequately compensate for cross-lingual and cross-center discrepancies. Among the fine-tuned models, we find that RoMistral-7B attains the highest resilience to both institutional and diagnostic variability. The fact that fine-tuned LLMs remain significantly above their zero-shot counterparts indicates that simple fine-tuning can bring useful question answering capabilities that generalize to different data distributions. Still, fine-tuning alone is not sufficient to achieve full generalization capacity across cancer types and medical centers, underlining the opportunity to introduce domain-adaptation strategies.

We further analyze the effect of prompt length. We test the models with various prompt lengths, taking into account the limitation imposed by each LLM architecture on its input prompt (see Tables \ref{tab:results} and~\ref{tab:secondary_results}). While the average length of an epicrisis is 7,829 tokens (as per Table \ref{tab:dataset_tokens}), the LLMs do not necessarily benefit from using extensive prompts. All of our models obtain superior performance levels with 2,048 input tokens than with 4,096 tokens or longer sequences.

Phi-4-mini-instruct is the only model that accepts very long prompts. Although we explore prompts of up to 32,768 tokens for Phi-4-mini-instruct, the model does not benefit from the extended input.

We further compare open-source vs.~API-based LLMs. To compare the open-source LLMs with state-of-the-art API-based models, we randomly select 100 QA pairs from the in-domain test set. The limited test set size is determined by our resource constraints, while the use of zero-shot prompting for the API-based models is due to the access restrictions imposed by OpenAI and Google, respectively. We present the comparative results in Table~\ref{tab:testset_random100_sota}. For a fair comparison, the results of each model are based on an optimal number of input tokens. GPT-5.2 and Gemini 3 Flash substantially outperform the lighter open-source models when both API-based and open-source LLMs are evaluated in the zero-shot setting. This highlights the effectiveness of large-scale general pretraining, even without task-specific supervision. Despite the strong performance of both GPT-5.2 and Gemini 3 Flash, all fine-tuned models consistently outperform the API-based LLMs across all evaluation metrics, confirming the benefits of task-specific and domain-specific fine-tuning. This comparison reinforces the importance of fine-tuning to achieve high performance on specialized clinical QA tasks in low-resource languages.

\begin{table*}[t!]
\centering
\caption{Results of Phi-4-mini-instruct on MedQARo, using only the first 2,048-token chunk of the epicrisis vs.~all 2,048-token chunks. The former approach significantly outperforms the latter one across all metrics, suggesting that only analyzing the first part of the epicrisis reduces input clutter and prevents overfitting. The best results are highlighted in \textcolor{darkgreen}{\textbf{bold green}}.}
\label{tab:alternative_eval}
\begin{adjustbox}{width=\textwidth}
\begin{tabular}{lc cccc c cccc}
\toprule
\multirow{2.5}{*}{\textbf{Model}} & \multirow{2.5}{*}{\textbf{Text chunks}} & \multicolumn{4}{c}{\textbf{Validation}} & & \multicolumn{4}{c}{\textbf{In-domain test}} \\
\cmidrule{3-6}
\cmidrule{8-11}
& & \textbf{F1} & \textbf{EM} & \textbf{BLEU} & \textbf{METEOR} & & \textbf{F1} & \textbf{EM} & \textbf{BLEU} & \textbf{METEOR} \\
\midrule
\multirow{2}{*}{\textbf{Phi-4-mini}} & First & \textcolor{darkgreen}{\textbf{0.6267}} &	\textcolor{darkgreen}{\textbf{0.6575}}	& \textcolor{darkgreen}{\textbf{0.5977}}	& \textcolor{darkgreen}{\textbf{0.3897}} &	& \textcolor{darkgreen}{\textbf{0.6291}}	& \textcolor{darkgreen}{\textbf{0.6593}} &	\textcolor{darkgreen}{\textbf{0.5999}} &	\textcolor{darkgreen}{\textbf{0.3916}} \\
 & All & 0.4354	& 0.4806	& 0.4162	& 0.2758	& & 0.4530 &	0.4966 &	0.4328	& 0.2839 \\
\bottomrule[0.8pt]
\end{tabular}
\end{adjustbox}
\end{table*}

We next assess the impact of using the first vs.~multiple input chunks.
Since epicrises are trimmed from the end to fit the desired prompt lengths, the generally low performance levels when using long prompts can be explained by two factors: (i) most questions can be answered just by looking at the beginning part of an epicrisis, or (ii) taking long contexts into account is a challenging learning task, potentially leading to overfitting. To distinguish between these two cases, we conduct another experiment with one of the best performing models, Phi-4-mini-instruct based on 2,048 input tokens. In addition to the version that simply trims the epicrisis from the end, we evaluate a version that analyzes all non-overlapping 2,048-token chunks from the epicrisis. For the latter approach, we employ majority voting over all chunks to obtain the final answer. To break ties, we rely on the average log-probability over the generated tokens and select the answer with the highest value. The corresponding results are reported in Table~\ref{tab:alternative_eval}. While the chunk-based approach enables processing of complete epicrises, it results in substantial performance degradation, suggesting that focusing on the first part of the epicrisis leads to optimal results. Still, as shown by our previous experiments reported in Table~\ref{tab:results}, using a context that is too short, e.g.~1,024 tokens for RoLLaMA2-7B-Instruct or RoMistral-7B-Instruct, might also lead to performance degradation. In summary, trimming the end part of the epicrisis seems to act as a regularization technique, and finding the optimal context length can be assimilated with the process of tuning the corresponding regularization hyperparameter.

% \todo{report and comment on the results of the best model (Phi-4-mini-instruct based on 3,072 tokens) for each of the three question categories.}

% Category           F1       EM     METEOR     BLEU
% --------------------------------------------------
% Binary          0.753    0.749      0.377    0.418
% Extractive      0.589    0.640      0.407    0.396
% Reasoning       0.572    0.569      0.323    0.509

\begin{figure*}[t]
  \centering
  \includegraphics[width=1.0\linewidth]{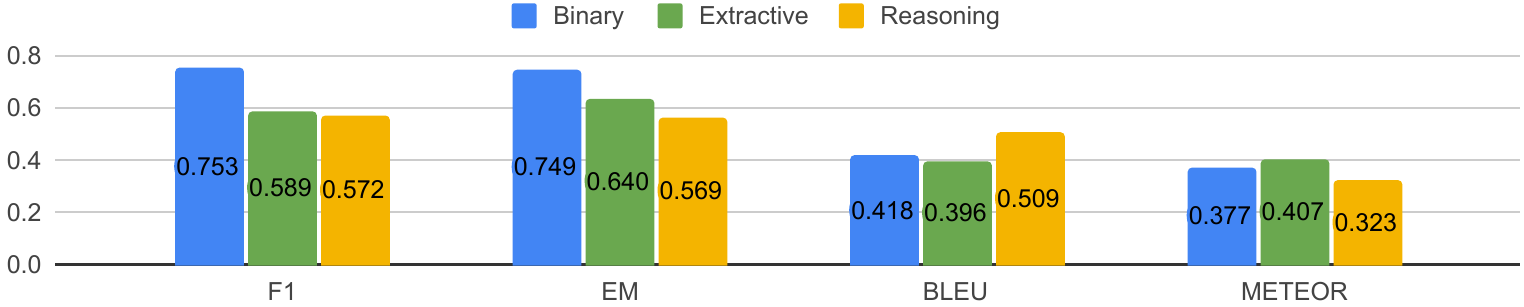}
  %\vspace{-0.4cm}
  \caption{Performance per question category (binary, extractive, and reasoning) across four evaluation metrics (F1, EM, BLEU, and METEOR). The values are reported for one of the top scoring models, namely Phi-4-mini-instruct based on 2,048 tokens, on the in-domain test set. Blue bars correspond to binary questions. Green bars correspond to extractive questions. Orange bars correspond to reasoning questions.}
  \label{fig:question-type-metrics}
\end{figure*}

We further showcase the importance of using multiple evaluation metrics.
Regarding the employed evaluation metrics, we observe that EM scores exceed F1 scores by small margins across most configurations, suggesting that the models tend to either match answers exactly or miss them entirely, with fewer partial matches. Moreover, METEOR scores are consistently lower than other metrics, likely reflecting the sensitivity of METEOR to synonyms and word variations in Romanian medical terminology, but also its inherent bias against short answers. Nevertheless, for a holistic evaluation of QA models on MedQARo, we promote the use of all four evaluation metrics.

We next analyze the performance per question type. In Figure~\ref{fig:question-type-metrics}, we present the results of one of our best performing model, for each of the three question categories. The F1 and EM metrics suggest that binary questions are easier to answer, while reasoning questions are harder. This observation is consistent with our natural expectation. However, the BLEU score seems to promote reasoning questions, while METEOR indicates that the model is more capable on extractive questions. Overall, the analysis shows that some metrics are more appropriate for certain question categories. This represents another strong argument in favor of using a wide variety of metrics to accurately benchmark models on MedQARo.

We further conclude our results with a number of general remarks. Our experimental results reveal several patterns across all models and configurations. First, we observe that in-domain validation and test results remain consistent across all experiments, indicating that our patient-level data splits effectively prevent data leakage. All fine-tuned LLMs exceed the naive baselines by large margins, confirming that models learn meaningful clinical question-answering capabilities. We also find that fine-tuned models significantly outperform their zero-shot counterparts across all architectures, emphasizing that neither prior Romanian language adaptation nor English biomedical pretraining is sufficient for the medical QA task. Among the four open-source LLM families, RoMistral-7B-Instruct obtains the highest in-domain performance levels. In contrast, OpenBio-LLM-8B achieves the lowest scores among all evaluated models on the in-domain test set, despite being pretrained on the medical domain. The cross-domain evaluation indicates that fine-tuning is also beneficial for patients enrolled in different medical centers, who suffer from other cancer types. However, fine-tuned LLMs exhibit a visible performance degradation in the cross-center and cross-diagnosis setup, suggesting that fine-tuning alone is not enough to recover the domain gap. Regarding the comparison between the zero-shot API-based LLMs and fine-tuned open-source LLMs, we find that GPT-5.2 and Gemini 3 Flash obtain reasonable performance levels, but they are clearly outperformed by fine-tuned LLMs. This finding emphasizes that fine-tuning open-source LLMs is preferred over zero-shot prompting larger models gated by paywalls.

\begin{table*}[ht]
\centering
\caption{Examples of test questions with correct and wrong answers provided by the best performing LLM, namely Phi-4-mini-instruct based on 3,072 tokens. Each row shows a question along with ground-truth and predicted answers. English translations are provided when necessary.}
\label{tab:MedQARo_qual}
\setlength{\tabcolsep}{3.8pt}
\small
\begin{tabular}{p{6.7cm} p{2.15cm} p{1.9cm} c}
\toprule
\textbf{Question (Romanian / English)} & \textbf{Ground-truth} & \textbf{Predicted} & \textbf{Correct?} \\
\midrule
% A fost necesar\u{a} ajustarea dozei tratamentului pentru a gestiona reac\c{t}ia advers\u{a} endocrin\u{a}? Was dose adjustment of the treatment necessary to manage the endocrine adverse event?	& Nu se aplica	& Nu se aplica & \textcolor{darkgreen}{\checkmark} \\
% \hline
Imunoterapia a fost introdus\u{a} \^{i}n a doua etap\u{a} a schemei terapeutice? / Was immunotherapy introduced in the second stage of the therapeutic regimen? & Nu / No & Nu / No & \textcolor{darkgreen}{\checkmark} \\
\midrule
S-au identificat metastaze la nivelul pl\u{a}m\^{a}nilor? / Are lung metastases present?	& Da / Yes & Da / Yes & \textcolor{darkgreen}{\checkmark} \\
\midrule
Care este stadiul tumoral la diagnostic al pacientului? / What was the tumor stage at the time of diagnosis for this patient? & IIA & IIA & \textcolor{darkgreen}{\checkmark} \\
\midrule
Care este subtipul histologic al pacientului din rezultatul histopatologic? / What is the histological subtype of the patient according to the histopathology report?	& ADK & ADK & \textcolor{darkgreen}{\checkmark} \\
% \hline
% Ce metod\u{a} de iradiere a fost aleas\u{a} pentru acest pacient? Which irradiation method was chosen for this patient? &	nu se stie	& nu se stie & \textcolor{darkgreen}{\checkmark} \\
\midrule
Ce cadrane ale s\^{a}nului sunt invadate de tumor\u{a}? / Which breast quadrants are invaded by the tumor? & supero-extern / upper-outer & supero-extern / upper-outer & \textcolor{darkgreen}{\checkmark} \\
\midrule
S-au utilizat alte linii de tratament \^{i}n progresia bolii? / Were other lines of treatment used during disease progression? & CHT	& Nu / No & \textcolor{red}{\xmark} \\
\midrule
S-au identificat metastaze la nivelul oaselor? / Are bone metastases present?	& Nu / No & Da / Yes & \textcolor{red}{\xmark} \\
\midrule
Sindrom inflamator este prezent la acest pacient? / Is the inflammatory syndrome present for this patient? & prezent / present	& absent / absent & \textcolor{red}{\xmark} \\
\midrule
Care este stadializarea TNM a pacientului? / What is the TNM staging of the patient? & cT4cT2cN2M0 / pT4bPNx	& cT4cN2M0 & \textcolor{red}{\xmark} \\
\midrule
Care este stadiul tumorii stabilit la \^{i}nceputul tratamentului? / What is the stage of the tumor established at the beginning of treatment? & IV & IIIB & \textcolor{red}{\xmark} \\
% \hline
% Cum a fost stabilit diagnosticul \^{i}n urma examin\u{a}rii clinice și a testelor de laborator? How was the diagnosis established following clinical examination and laboratory tests? & Neoplasm mamar stang, (std IIA T1cN1M0) biopsiat (receptori hormonali poz, Her2 neg), in curs de HT (01.2018  - Anastrozol 1mg), operata (09.2018 mastectomie radicala stg modificata tip Madden) stadiu pT2N0Mx & Neoplasm mamar stang,  (std & \\
\bottomrule[0.8pt]
\end{tabular}
\end{table*}

\vspace{-0.1cm}
\subsection*{Qualitative Analysis}
\vspace{-0.1cm}

To better understand the challenges posed by MedQARo and the capabilities of one of the best-performing model, we conduct a qualitative analysis of a random selection of QA pairs with correct and wrong answers predicted by Phi-4-mini-instruct. In Table \ref{tab:MedQARo_qual}, we showcase 10 examples, 5 that are correctly answered and 5 that are incorrectly answered, respectively. 

We start by analyzing correctly answered questions. In the first example, the LLM was able to recognize the sequence of treatment lines and to distinguish between the first line and the second line of therapy, even though the information was not explicitly stated in the epicrisis. In the second example, pulmonary metastases were mentioned under the abbreviation M1PUL. The model recognized that M1PUL denotes the presence of lung metastases, providing the correct answer based on this information. In the third example, the LLM demonstrates the ability to correctly stage breast cancer, which requires high-level medical reasoning. The model correlated tumor size, regional invasion, and the absence of distant metastases, successfully assigning stage IIA, according to the TNM criteria. In the fourth case, the histological type ``adenocarcinom'' / ``adenocarcinoma'' was explicitly stated in the histopathology report, which allowed the LLM to correctly answer with ``ADK'' to the posed question. In the fifth example, the interpretation of the chest CT explicitly stated ``invazia tumoral\u{a} \^{i}n cadranul supero-extern'' / ``tumor invasion present in the upper-outer quadrant''. The model was able to accurately spot this information and provide the correct answer. Overall, we observe that the model demonstrates the necessary capabilities to answer both reasoning and extractive questions. 

We continue by analyzing wrongly answered questions. In the first example, the annotator considered that the presence of ``CHT'' in the epicrisis indicates a new line of treatment at disease progression. However, the LLM answered ``Nu'' / ``No'', either because it did not recognize that ``CHT'' represented a new line of therapy, or because it required an explicit statement confirming the change of treatment line, which is missing from the epicrisis. In the latter case, the model might have considered that ``CHT'' is part of the initial treatment, instead of a new line of therapy. Note that this is not exactly a binary question, since the correct answer for this kind of question is a list of treatment lines. When the list is empty, the correct answer is ``Nu'' / ``No''. In the second example, the LLM may have been triggered by the presence of the word ``os'' / ``bone''. The confusion is caused by an investigation reported in the epicrisis, called ``scintigrafie osoas\u{a}'' / ``bone scintigraphy'', which is usually recommended for patients with bone metastases. Despite undergoing this investigation, the patient did not have any bone metastases. In the third case, the inflammatory syndrome is not directly mentioned in the epicrisis, although it can be identified by analyzing biological parameters (e.g.~elevated ESR, increased CRP, elevated fibrinogen) or secondary diagnoses (e.g.~infection, chronic inflammation). The physician recognized these signs and determined the presence of the inflammatory syndrome. Without an explicit mention of ``sindrom inflamator'' / ``inflammatory syndrome'', the LLM was not able to extract the correct answer. In the fourth case, the LLM provided only a partially correct answer. It correctly identified the clinical TNM classification, but did not mention the pathological TNM classification. This is a limitation induced by the prompt length of 3,072 tokens, since the pathological TNM classification is usually specified near the end of the epicrisis, which gets removed due to context trimming. In the fifth example, the patient had lymph node metastases. The physician considered the staging to be stage IV, whereas the LLM interpreted the involvement as loco-regional lymph node invasion and classified the case as stage IIIB. Overall, the wrong or partially incorrect answers point out two kinds of limitations: (i) the reasoning abilities of Phi-4-mini-instruct can sometimes hinder the integration of high-level medical information, and (ii) the context trimming procedure might sometimes lead to dropping important clues.

\vspace{-0.1cm}
\section*{Discussion}
\vspace{-0.1cm}

In this paper, we introduced MedQARo, the first large-scale benchmark for Romanian medical question answering. The new benchmark is the result of a meticulous data collection and annotation process carried out by experts in the medical domain, who constructed a dataset of 105,880 QA pairs. The QA samples are grounded in a collection of 1,242 medical case summaries of cancer patients. We conducted comprehensive experiments with several LLM families to assess the impact of various aspects, namely prompt structure and length, language vs.~domain vs.~general pretraining, and zero-shot prompting vs.~LoRA-based fine-tuning. Our empirical results suggested some interesting findings. First of all, we observed that long prompts are not always helpful, degrading performance even if epicrises are typically long documents. Second of all, we found that language specialization (e.g.~RoMistral-7B) and generic pretraining (e.g.~Phi-4-mini) offer a better starting point for fine-tuning than medical domain specialization (e.g.~OpenBioLLM-8B). Third of all, we determined that fine-tuning on our specific task significantly outperforms zero-shot prompting. Yet, even the best model was only able to correctly answer less than $70\%$ of the in-domain test questions and $40\%$ of the cross-domain test questions, indicating that MedQARo is a challenging benchmark.

In future work, we aim to develop models that integrate retrieval-augmented generation to improve question answering capabilities. More specifically, we will focus on developing a retrieval module to accurately find the sections of the input epicrisis that are likely to contain indicators for the answer. This will allow us to use models with reasonable context lengths, while minimizing the risk of eliminating important information during context trimming. However, since the answer is not always directly mentioned in the epicrisis, finding the right sections from which the answer could be derived requires careful consideration. 

\section*{Methods}

%\vspace{-0.1cm}
\subsection*{Data Collection}
%\vspace{-0.1cm}

%\todo{Andreea: De scris despre procesul de colectare / extragere a datelor pentru cele doua seturi de date. Cati doctori au participat? Cum au fost verificate datele extrase? Care a fost numarul minim / maxim de pacienti per doctor? Ce reguli ati stabilit pentru extragere (conventii legate de utilizare denumiri medicale, prescurtare, etc.)? Detalii despre consimtamantul pacientilor pentr a le utiliza datele in scop de research. Orice altceva nu am spus eu mai devreme. Totul in limba engleza.}

MedQARo is obtained through the manual extraction of information from medical documents, namely medical summaries of patients that are registered either in the Oncology Department of Col\c{t}ea Clinical Hospital, or in the ``Dan Furtun\u{a}'' Medical Center, both located in Bucharest, Romania. The epicrises, originally drafted in Word format by the physicians attending each patient, were manually reviewed by a group of seven physicians involved in constructing MedQARo. In this group, two experienced physicians from Col\c{t}ea Clinical Hospital, who led the data collection process, established the most prevalent cancer sites in their hospital, namely breast and lung. They selected patients with either one of these diagnoses, who received treatment at the hospital between 2019 and 2024. This selection process led to a set of 796 patients with breast cancer and 215 patients with lung cancer. To create a diverse multi-center and multi-cancer benchmark, 231 patients with one of five cancer types (head and neck, melanoma, renal cell carcinoma, bladder cancer, hepatocellular carcinoma) were included in the study. These patients are enrolled in cancer follow-up care programs at ``Dan Furtun\u{a}'' Medical Center.

The two experienced physicians further proposed a set of reference questions for each cancer site, comprising 48 questions for breast cancer, 61 questions for lung cancer, and 7 general questions that cover all other types. The other five physicians proposed 8 to 20 equivalent reformulations of the reference questions. Next, the physicians proceeded by extracting the answers for each unique question from each epicrisis and entering the respective answers into a database. To optimize annotation time, the questions for each epicrisis are grouped and annotated in the same batch. This means that a physician could read an epicrisis once, before answering multiple questions about the respective patient.

Data extraction and completion were carried out via a two-level process. The first level was performed by five medical oncology and radiotherapy residents, who meticulously completed the database of QA pairs following prior training based on the annotation guidelines provided by the experienced physicians. The second level was carried out by two board-certified medical oncology specialists, holders of a good standing certificate issued by the Romanian College of Physicians and with proven experience in good medical and clinical research practices. They verified the accuracy and consistency of the data entered by the residents. In total, the annotation process involved seven physicians, with a cumulative effort of approximately 2,170 hours (equivalent to about 310 hours per physician) just to type in the answers. Neither the time required to formulate questions nor the time required to read an epicrisis before answering all questions for a certain patient is taken into account in the time measurement. While the annotation interface did not allow us to measure the extra time required to read epicrises, we estimate that this step required between 600 and 800 additional hours. The time required to formulate all question variants is estimated at 50-70 hours. Adding all numbers, we estimate that the entire annotation process took nearly 3,000 hours. The number of patients processed by each resident ranged from a minimum of 200 to a maximum of 280 cases, depending on individual availability and case complexity. To measure the inter-annotator agreement, a number of 350 randomly chosen QA pairs were annotated by all five physicians. The average Cohen's $\kappa$ score between each pair of annotators is $0.9562$, which indicates an almost perfect agreement.

Furthermore, each experienced specialist meticulously reviewed the QA pairs associated with about 600-630 patients. This distribution of the workload was planned to ensure a relatively balanced annotation process. For data extraction, the experienced physicians developed an annotation guideline, establishing clear completion rules, including: staging of cancers according to the Tumor Node Metastasis (TNM) classification; grading of adverse events according to the Common Terminology Criteria for Adverse Events (CTCAE) system; standardization of medical abbreviations (e.g.~``CHT'' for chemotherapy, ``RT'' for radiotherapy, ``NK'' for ``not known'' values); handling incomplete information. In addition to these general rules, the guideline included a detailed list of answer options to be extracted and recorded in a standardized format: Eastern Cooperative Oncology Group (ECOG) performance status, menopausal status, molecular subtype, tumor proliferation index (Ki-67), histological grade, tumor size, family history of cancer, personal history of other cancers, body mass index (BMI), tumor markers, presence/absence of disease progression, number of metastatic sites, metastatic locations, death, etc. The senior physicians also acted as referees to solve annotation conflicts, choosing the final answer between two options based on their own high-skilled assessment.

\subsection*{Ethics Approval}

Regarding research ethics, we emphasize that all data was fully anonymized prior to processing by removing any direct identifiers (names, dates of birth, personal identification numbers, etc.). The use of anonymized data for research purposes is based on the informed consent of patients, as recorded in the medical charts of Col\c{t}ea Clinical Hospital and ``Dan Furtun\u{a}'' Medical Center, and signed individually by each patient at admission. The database was approved by the Ethics Committee of Col\c{t}ea Clinical Hospital (Anca Lupu, Ciprian Aldea, Traian Burco\c{s}, \c{S}erban Berte\c{s}teanu, Lavinia Co\c{t}ofan\u{a}, Leti\c{t}ia Coriu, Gabriela Caragescu, Livia Neac\c{s}u) where the research was conducted, in accordance with Decision No.~19091, dated October 5th, 2021. The database was also approved by the Ethics Committee of ``Dan Furtun\u{a}'' Medical Center (Bogdan Ioan Furtun\u{a}, Cipriana Dimitrakopoulos, Alice Corina F\u{a}rca\c{s}, Natalia B\u{a}rcaru, Daniela Tomescu), as per Decision No.~1323, dated 11th December, 2025. The entire process complies with the requirements of the General Data Protection Regulation (GDPR), the principles of the Declaration of Helsinki, and the standards of good medical and research practices.

\begin{table*}[t!]
\centering
\caption{A summary of the number of patients and corresponding QA pairs included in the MedQARo benchmark across cancer types. For breast and lung cancer, the splits are performed at the patient level, into training (70\%), validation (15\%), and in-domain test (15\%), to prevent information leakage across splits. A cross-domain test set comprises a separate set of 231 patients with one of five cancer types (excluding breast and lung), from a different medical center. We report the number of patients per cancer type per split, and the corresponding number of QA samples generated for each split. The final row aggregates the total number of unique patients and QA pairs across different cancer types for each split. MedQARo comprises a total 105,880 QA pairs for 1,242 patients.}
\label{tab:dataset_patients}
\resizebox{\textwidth}{!}{%
\setlength\tabcolsep{3pt}
\begin{tabular}{lcccc c cccc}
\toprule
\multirow{2.5}{*}{\textbf{Patient group}} & \multicolumn{4}{c}{\textbf{Number of patients}} & & \multicolumn{4}{c}{\textbf{Number of samples}} \\
\cmidrule{2-5}
\cmidrule{7-10}
& \textbf{Train} & \textbf{Validation} & \textbf{Test} & \textbf{Total} & & \textbf{Train} & \textbf{Validation} & \textbf{Test} & \textbf{Total} \\
\midrule
Breast cancer & 557 & 119 & 120 & 796 & & 53,472 & 11,424 & 11,520 & 76,416 \\
Lung cancer & 150 & 32 & 33 & 215 & & 18,300 & 3,904 & 4,026 & 26,230 \\
Other cancers & - & - & 231 & 231 & & - & - & 3,234 & 3,234  \\
\midrule
Total & 707 & 151 & 384 & 1,242 & & 71,772 & 15,328 & 18,780 & 105,880 \\
\bottomrule[0.8pt]%
\end{tabular}%
}
\end{table*}

%\vspace{-0.1cm}
\subsection*{Data Preprocessing}
%\vspace{-0.1cm}

To support model training and evaluation, we constructed two domain-specific datasets targeting question answering in oncology given a clinical context: one for breast cancer and one for lung cancer. These are complemented by a generic dataset targeting generic questions covering five other cancer types (not including breast and lung cancer).

% \noindent
% \textbf{Breast cancer.}
We collected data from 796 breast cancer patients, each associated with a complete epicrisis, i.e.~a detailed clinical summary for each patient, and a corresponding set of answers to 48 unique medical questions. To simulate natural variation in clinical queries, each question is reformulated by multiple physicians, leading multiple alternative formulations. For every patient-question pair, we randomly sample two question alternatives and generate two QA instances, each structured as a tuple of the form (patient ID, epicrisis, question, answer). This process resulted in a total of 76,416 unique examples.

% \noindent
% \textbf{Lung cancer.}
We compiled data from another 215 lung patients, each accompanied by an epicrisis summarizing their clinical history. For each patient, we curated a set of answers to 61 unique and clinically relevant questions. Similar to the breast cancer setup, each question is manually paraphrased into multiple linguistically diverse formulations to better reflect natural question variability. For each patient-question pair, we randomly sampled two different question variants. This leads to a total of 26,230 unique QA instances, capturing both clinical diversity and linguistic variation.

% \noindent
% \textbf{Other cancers.} 
To construct a benchmark that enables the assessment of generalization capacity across medical centers and cancer types, we created a secondary dataset, which is not used during training. This dataset comprises 231 patients, diagnosed with a different kind of cancer (head and neck, melanoma, renal cell carcinoma, bladder cancer, or hepatocellular carcinoma), all coming from a different center than the patients included in our main dataset. For these patients, there are 7 general questions, relevant for all five cancer types. As for breast and lung cancer patients, we randomly sampled two different question variants for each patient-question pair, leading to a total of 3,234 unique QA pairs.

% Breast data
% Bin range	Count		Gender	Count
% 0–4	0		F	796
% 5–9	0		M	0
% 10–14	0			
% 15–19	0			
% 20–24	3			
% 25–29	1			
% 30–34	8			
% 35–39	14			
% 40–44	36			
% 45–49	49			
% 50–54	83			
% 55–59	68			
% 60–64	109			
% 65–69	162			
% 70–74	130			
% 75–79	80			
% 80–84	40			
% 85–89	12			
% 90–94	1			
% 95–99	0		

% Lung data
% Bin range	Count			Gender	Count
% 0–4	0			F	66
% 5–9	0			M	149
% 10–14	0				
% 15–19	0				
% 20–24	0				
% 25–29	0				
% 30–34	1				
% 35–39	3				
% 40–44	1				
% 45–49	4				
% 50–54	25				
% 55–59	27				
% 60–64	37				
% 65–69	59				
% 70–74	39				
% 75–79	16				
% 80–84	3				
% 85–89	0				
% 90–94	0				
% 95–99	0				

% Other diagnosis
% Bin range	Count			Gender	Count		Diagnosis	Count
% 0–4	0			F	54		ORL	147
% 5–9	0			M	177		MELANOM	49
% 10–14	0						RCC	18
% 15–19	0						UROTELIAL	13
% 20–24	0						HCC	4
% 25–29	0							
% 30–34	4							
% 35–39	3							
% 40–44	8							
% 45–49	26							
% 50–54	29							
% 55–59	26							
% 60–64	37							
% 65–69	50							
% 70–74	26							
% 75–79	13							
% 80–84	7							
% 85–89	1							
% 90–94	1							
% 95–99	0							

%\vspace{-0.1cm}

%\vspace{-0.1cm}

\subsection*{Data Partitioning and Statistics} 
%\vspace{-0.1cm}

We merge the breast and lung cancer patients (from the Col\c{t}ea Clinical Hospital) and corresponding QA pairs into a single dataset. Then, the merged dataset is split into training, validation and in-domain test, using a $70\%/15\%/15\%$ split ratio, ensuring that the division occurs at the patient level. We prefer this split over splitting individual QA pairs because it guarantees that all the data from a single patient, including their epicrisis and associated QA pairs, remains within exactly one subset, thereby preventing data leakage among splits. Furthermore, we keep the patients with other cancers (from the ``Dan Furtun\u{a}'' Medical Center) for cross-domain evaluation.

\begin{table}[t!]
\centering
\caption{A summary of the minimum, maximum, average and total number of tokens for the three kinds of data (epicrisis, question and answer) available in MedQARo.}
\label{tab:dataset_tokens}
\begin{tabular}{lcccc}
\toprule
\multirow{2.5}{*}{\textbf{Data type}} & \multicolumn{4}{c}{\textbf{Number of tokens}} \\
\cmidrule{2-5}
 & \textbf{Min} & \textbf{Max} & \textbf{Mean} & \textbf{Total} \\
\midrule
% Epicrisis & 207 & 34,247 & 7,171 & 7,249,881 \\
Epicrisis & 207 & 34,247 & 7,829 & 9,723,618 \\
Question & 6 & 32 & 16 & 1,778,675 \\
Answer & 1 & 442 & 4 & 432,754 \\
\bottomrule[0.9pt]
\end{tabular}
\end{table}

We present the distribution of our dataset across training, validation, and test sets in Table~\ref{tab:dataset_patients}. For the breast cancer subset, our split results in 557 patients allocated for training, 119 for validation, and 120 for testing. In terms of QA pairs, the split distributes 53,472 samples for training, 11,424 for validation, and 11,520 for testing, respectively. Similarly, the lung cancer subset is divided into 150 training, 32 validation, and 33 test patients, yielding 18,300, 3,904, and 4,026 QA pairs, respectively. A separate cross-domain test set contains 231 patients and 3,234 QA pairs. In summary, MedQARo contains four data partitions: one for training (707 patients), one for in-domain validation (151 patients), one for in-domain testing (153 patients), and one for cross-domain testing (231 patients). In total, MedQARo comprises 1,242 unique patients (796 suffering from breast cancer, 215 from lung cancer, and 231 from five other cancers) and 105,880 question-answer pairs. Hence, our contribution is a substantial resource for training and evaluating Romanian medical QA systems, in both in-domain and cross-domain scenarios.

% \todo{specify min, max and average number of tokens per epicrisis, per question, per answer in Table \ref{tab:dataset_tokens}. add comments for the table}
In Table~\ref{tab:dataset_tokens}, we report statistics about the number of tokens in epicrises, questions and answers included in MedQARo. Epicrises range from 207 to 34,247 tokens, with a mean value of 7,829 tokens, where shorter documents typically correspond to patients under investigation or at early treatment stages, while longer ones capture extensive treatment histories with multiple interventions. This variation can pose challenges for some models with limited context windows. While epicrises are typically lengthy, questions and answers are much shorter. Questions maintain relative consistency, between 7 and 30 tokens, with a mean value of 15 tokens, reflecting the consistency of clinical query formulations. The answers are represented by predominantly brief texts with a mean of 4 tokens, varying in length from one to 442 tokens. Longer responses are typical for questions involving complex medical reasoning.

\begin{table}[t]
\centering
\caption{Representative examples of binary, extractive, and reasoning questions from MedQARo. Each question is shown in Romanian, followed by its English translation. One reformulation is displayed below each distinct question.}
\label{tab:MedQARo_examples}
\setlength{\tabcolsep}{4pt}
\begin{tabular}{p{1.5cm} p{11.1cm}}
\toprule
\textbf{Type} & \textbf{Question (Romanian / English)} \\
\midrule
\multirow{11}{*}{Binary} &
S-a realizat o interven\c{t}ie chirurgical\u{a} \^{i}nainte de ini\c{t}ierea imunoterapiei? / Has a surgical intervention been performed before initiating immunotherapy? \\
& Interven\c{t}ia chirurgical\u{a} a precedat imunoterapia \^{i}n cazul acestui pacient? / Did surgery precede immunotherapy for this patient? \\
\cmidrule{2-2}
& Pacientul are gena BRCA1? / Does the patient have the BRCA1 gene? \\
& Rezultatul analizei genetice indic\u{a} prezen\c{t}a muta\c{t}iei BRCA1? / Does the genetic analysis result indicate the presence of the BRCA1 mutation? \\
\cmidrule{2-2}
& S-au identificat metastaze la nivelul pl\u{a}m\^{a}nilor? / Have lung metastases been identified? \\
& Exist\u{a} determin\u{a}ri secundare pulmonare la acest pacient? / Are there any secondary pulmonary tumors in this patient? \\

\midrule
\multirow{8}{*}{Extractive} &
Care a fost dimensiunea tumorii la debut? / What was the tumor size at the time of diagnosis? \\
& C\^{a}t de mare era tumora atunci c\^{a}nd a fost stabilit diagnosticul? / How large was the tumor when the diagnosis was made? \\
\cmidrule{2-2}
& Ce cadrane ale s\^{a}nului sunt invadate de tumor\u{a}? / Which breast quadrants are invaded by the tumor? \\
& Specific\u{a} cadranele invadate de tumor\u{a} / Specify the quadrants invaded by the tumor \\
\cmidrule{2-2}
& Ce status ECOG are pacientul? / What is the ECOG status of the pacient? \\
& Specific\u{a} statusul de performan\c{t}\u{a} ECOG / Specify the ECOG performance status \\
\midrule
\multirow{12}{*}{Reasoning} &
C\^{a}te luni sunt de la \^{i}nceputul tratamentului p\^{a}n\u{a} la apari\c{t}ia progresiei bolii? / How many months have passed from the start of treatment until the onset of disease progression? \\
& C\^{a}t a durat (\^{i}n luni) ca boala s\u{a} progreseze sub tratament? / How long did it take (in months) for the disease to progress under treatment? \\
\cmidrule{2-2}
& C\^{a}te luni au trecut de la chirurgie p\^{a}n\u{a} la \^{i}nceperea tratamentului cu radioterapie? / How many months have elapsed from surgery until the initiation of radiotherapy treatment? \\
& La c\^{a}t timp dup\u{a} opera\c{t}ie a \^{i}nceput tratamentul radioterapic? / How long after surgery did radiotherapy treatment begin? \\
\cmidrule{2-2}
& Care este stadiul tumoral la diagnostic al pacientului? / What was the tumor stage at the time of diagnosis for this patient? \\
& Ce stadiu de cancer a fost stabilit la diagnostic? / What stage of cancer was determined at diagnosis? \\
\bottomrule[0.8pt]
\end{tabular}
\end{table}

\begin{figure}[t]
    \centering
    \includegraphics[width=0.5\linewidth]{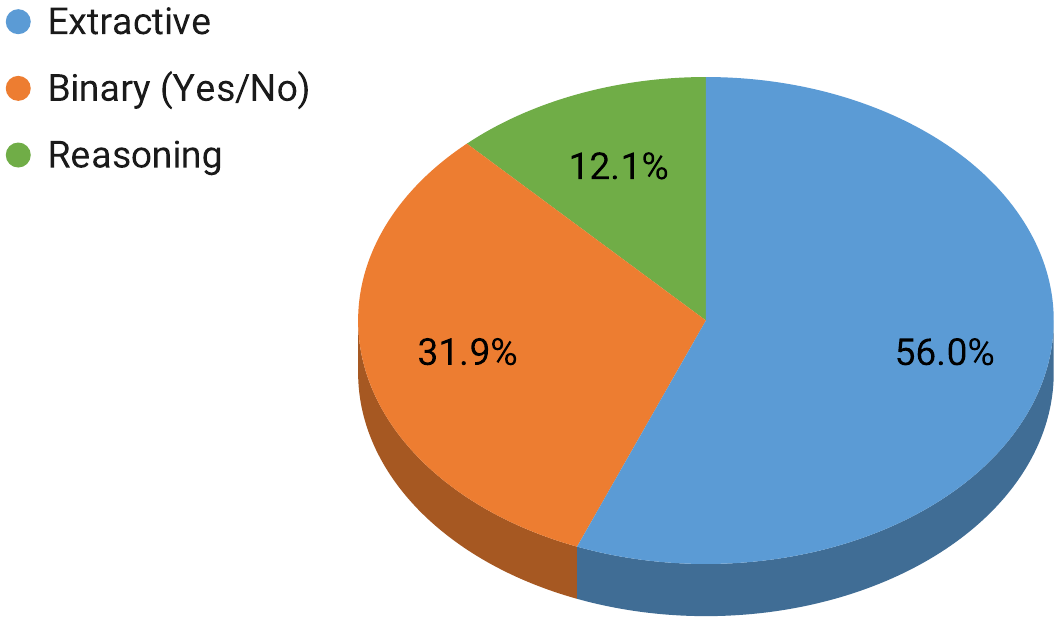}
    \caption{Distribution of question types in the MedQARo dataset. The dataset comprises three main categories: binary (yes/no) questions, 
    extractive questions (answers explicitly found in the epicrisis), 
    and reasoning questions (requiring inference beyond explicit mentions). Percentages are computed over the full set of 105{,}880 QA pairs. Best viewed in color. Blue represents extractive questions. Orange represents binary questions. Green represents reasoning questions.}
    \label{fig:MedQARo_question_types}
    %\vspace{-0.2cm}
\end{figure}

We categorize the questions into three types: (i) binary (yes/no); (ii) extractive, where the answer appears explicitly in the clinical context (epicrisis); and (iii) reasoning, where the answer must be inferred from multiple clues within the epicrisis. The distribution of these categories in MedQARo is shown in Figure~\ref{fig:MedQARo_question_types}. Within the binary subset, the label distribution is moderately skewed, with roughly two thirds of answers being ``Nu'' (``No'') and one third being ``Da'' (``Yes''). The slightly skewed distribution resulted naturally from the data collection process. In Table~\ref{tab:MedQARo_examples}, we showcase some representative examples from the three question categories included in MedQARo. Some questions are specific to a certain cancer type, e.g. ``Ce cadrane ale s\^{a}nului sunt invadate de tumor\u{a}?'' / ``Which breast quadrants are invaded by the tumor?'' is only relevant for breast cancer, while other questions are valid for all cancer types, e.g. ``Care este stadiul tumoral la diagnostic al pacientului?'' / ``What was the tumor stage at the time of diagnosis for this patient?'' is applicable to all cancers. The reformulated questions indicate that semantic equivalence was ensured during the process of manual question generation conducted by the annotators.

\begin{figure*}[!t]
% \begin{subfigure}[t]{.54\textwidth}
%   \centering
%     \caption{}
%       \label{fig:breast_age}
%   \includegraphics[width=1.0\linewidth]{fig_breast_age.pdf} 
%   % \vspace{-0.3cm}
% \end{subfigure}
% \hfill
% \begin{subfigure}[t]{.43\textwidth}
%   \centering
%   \caption{}
%   \label{fig:breast_gender}
%   \includegraphics[width=1.0\linewidth]{fig_breast_gender.pdf}  
%   % \vspace{-0.3cm}
% \end{subfigure}\\
% \begin{subfigure}[t]{.54\textwidth}
%   \centering
%   \caption{}
%   \label{fig:lung_age}
% \includegraphics[width=1.0\textwidth]{fig_lung_age.pdf}  
% % \vspace{-0.3cm}
% \end{subfigure}
% \hfill
% \begin{subfigure}[t]{.43\textwidth}
%   \centering
%   \caption{}
%   \label{fig:lung_gender}
% \includegraphics[width=1.0\textwidth]{fig_lung_gender.pdf}  
% %\vspace{-0.3cm}
% \end{subfigure}\\
% \begin{subfigure}[t]{.54\textwidth}
%   \centering
%   \caption{}
%   \label{fig:other_age}
% \includegraphics[width=1.0\textwidth]{fig_other_age.pdf}  
% % \vspace{-0.3cm}
% \end{subfigure}
% \hfill
% \begin{subfigure}[t]{.43\textwidth}
% \caption{}
%   \label{fig:other_gender}
%   \centering
% \includegraphics[width=1.0\textwidth]{fig_other_gender.pdf} 
% %\vspace{-0.3cm}
% \end{subfigure}
\includegraphics[width=1.0\linewidth]{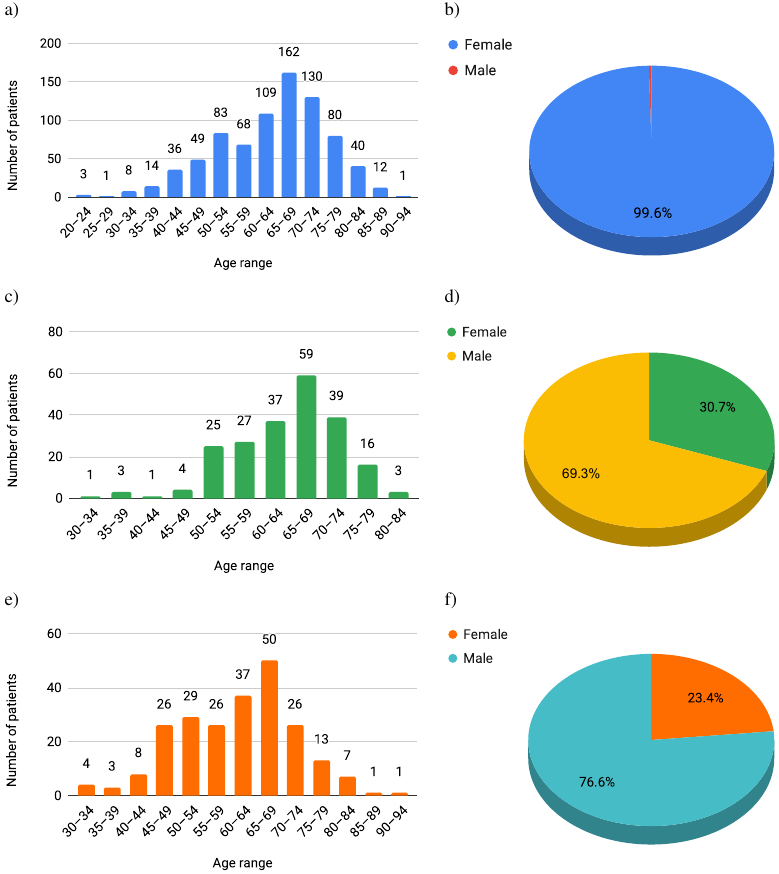}
\caption{Age and gender distributions of patients included in MedQARo. Separate distributions are reported for breast, lung, and other cancer patients. Age distributions are based on 5-year bins (between 20 and 94 years). a) Age distribution for breast cancer. b) Gender distribution for breast cancer. c) Age distribution for lung cancer. d) Gender distribution for lung cancer. e) Age distribution for other cancers. f) Gender distribution for other cancers.}
\label{fig:demographics}
\end{figure*}

\begin{figure}
    \centering
    \includegraphics[width=0.65\textwidth]{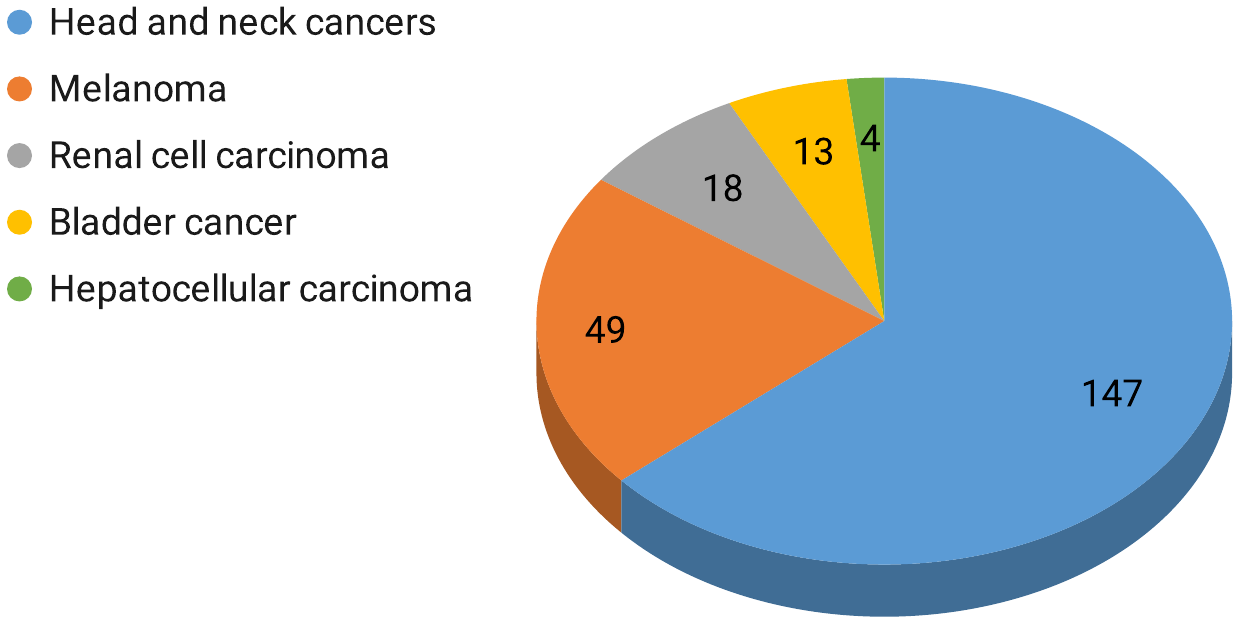}
    \caption{Distribution across cancer types for patients included in the cross-domain test set of MedQARo. Blue represents head and neck cancers. Orange represents melanoma. Gray represents renal cell carcinoma. Yellow represents bladder cancer. Green represents hepatocellular carcinoma.}
    \label{fig:other_diagnoses_demographics}
\end{figure}

\subsection*{Data Demographics}

We further examine the age and gender distributions of patients included in MedQARo. In Figure~\ref{fig:demographics}, we report the distributions for the three patient groups, which are separated by cancer type: breast cancer, lung cancer, and other cancers. Across all diagnostic categories, the cohort is predominantly composed of middle-aged and elderly patients, with most cases falling in the 50-79 age range. As naturally expected, most breast cancer patients are females (793 out of 796), as breast cancer is less common in the male population. In contrast, the male population is more affected by lung cancer and other cancers, which can be attributed to the often negligent behavior of males towards maintaining a healthy lifestyle \cite{Bonhomme-JMHG-2007,Dong-G-2012}.

In Figure~\ref{fig:other_diagnoses_demographics}, we illustrate the distribution of patients diagnosed with other cancer types. The distribution reveals heterogeneous demographic patterns across diagnoses, with head and neck cancers accounting for the largest proportion of cases, followed by melanoma.

These demographic patterns are consistent with real-world clinical populations and suggest that MedQARo captures representative patient characteristics, supporting the etiological validity of the benchmark for clinical question answering.

%\vspace{-0.1cm}
\subsection*{Overview of Large Language Models}
%\vspace{-0.1cm}

To address the task of medical question answering in Romanian, we consider a diverse set of LLMs and use them under various configurations, considering both zero-shot and fine-tuning settings. Our goal is to understand how distinct architectural specializations (e.g.~language-specific, domain-specific, or long-context models) and alternative prompt formulations affect performance on the Romanian clinical QA task. We select six distinct LLM families. Two of them are pretrained on Romanian, namely RoLLaMA2-7B \citep{masala2024vorbecstiromanecsterecipetrain}, which is adapted from LLaMA 2 \citep{touvron2023llama}, and RoMistral-7B \citep{masala2024vorbecstiromanecsterecipetrain}, which is adapted from the Mistral architecture \citep{jiang2023mistral7b}. The third model, namely Phi-4-mini-instruct \citep{phi4mini}, is optimized for long-context inputs. This is useful for the processing of entire epicrises, which can often get to be very long (see Table \ref{tab:dataset_tokens}), especially for patients with a long medical history. Given the target domain, the fourth model, LLaMA3-OpenBioLLM-8B \citep{OpenBioLLMs}, is pretrained on biomedical corpora. Finally, we evaluate two state-of-the-art LLMs, GPT-5.2 \cite{GPT52-2025} and Gemini 3 Flash \cite{Gemini3-2025}, that are only accessible via public APIs.

The first four LLMs are employed in two setups, zero-shot prompting and end-to-end fine-tuning. Due to the large number of parameters in the chosen models, we adopt a parameter-efficient fine-tuning strategy based on Low-Rank Adaptation (LoRA) \citep{hu2021lora} specialized on the causal language modeling task. By comparing the results obtained in the two setups, we can precisely contextualize the impact of fine-tuning on MedQARo. In addition, we also test two alternative prompts, where the order between context (epicrisis) and question is interchanged. To determine the impact of zero-shot prompting, we include a random token selector and a majority answer model as baselines for the QA task. Since access to the architectures and weights of GPT-5.2 and Gemini 3 Flash is not available, we only employ these two models in the zero-shot setup. Next, we describe all evaluated methods in detail.

%\vspace{-0.1cm}
%\vspace{-0.1cm}

\subsection*{Romanian-Adapted LLMs}

We start from two instruction-tuned language models specifically adapted for Romanian, RoLLaMA2-7B-Instruct and RoMistral-7B-Instruct. Both models are part of the recent efforts to build high-quality Romanian LLMs by further pretraining on Romanian corpora and applying instruction tuning with Romanian datasets, such as RoAlpaca \citep{alpaca, masala2024vorbecstiromanecsterecipetrain}, RoOpenAssistant \citep{kopf2024openassistant, masala2024vorbecstiromanecsterecipetrain} and RoOrca \citep{mukherjee2023orca, masala2024vorbecstiromanecsterecipetrain}. The two LLMs have been optimized for Romanian morphology and syntax using tokenizers adapted to the language. They have shown strong performance on Romanian natural language understanding and generation tasks \citep{masala2024vorbecstiromanecsterecipetrain}. In our study, we further fine-tune both models on MedQARo and compare their performance under different prompt structures and lengths.

As it inherits the architectural constraints of the original LLaMA 2 model \citep{touvron2023llama}, RoLLaMA2-7B is limited to a context window of 4,096 tokens. We employ a LoRA configuration specialized on the causal language modeling task. The chosen configuration enables updating a small subset of the model parameters ($0.0622\%$) during fine-tuning, significantly reducing training overhead. To enhance the adaptability of the LLM to the clinical QA task, we further consider several configuration alternatives. First, we evaluate on various prompt lengths, considering limits of 1,024, 2,048 and 4,096 tokens. If an epicrisis is longer than the accepted input length, we simply trim it from the end. When evaluating on prompts of 2,048 tokens, we also explore two prompt formats: \texttt{Q+E+A} (question + epicrisis + answer) and \texttt{E+Q+A} (epicrisis + question + answer). The model is trained via token-level cross-entropy, where the task is to predict the answer tokens via next token prediction. For this reason, the answer is always placed at the end of the prompt.

% To train the model, we first compile a list of unique answers from the entire training set and transform them into classes. This enables us to train the model via cross-entropy, where the number of classes (unique answers) is \todo{XXX}.

% Similarly to the base Mistral model, RoMistral-7B-Instruct is designed for extended input lengths, supporting up to 8,192 tokens. 
Given the large number of parameters in the base Mistral model, we again employ a LoRA-based fine-tuning strategy, enabling updates to only a small number of parameters. In this case, we fine-tune just $0.0470\%$ of the parameters, using the same LoRA configuration as for RoLLaMA2-7B-Instruct. This approach ensures both memory efficiency and effective domain adaptation to the Romanian clinical QA setting. We conduct experiments with two prompt lengths, namely 2,048 and 4,096. Consistent with the preliminary empirical results obtained with the previous model, the prompt format is set to \texttt{Q+E+A} in all the experiments performed with RoMistral-7B-Instruct. We employ token-level cross-entropy to optimize the model.

\subsection*{Long-Context LLM}

Phi-4-mini-instruct is a compact instruction-tuned transformer model designed for long-context understanding, offering efficient inference for up to 128K tokens. We employ this model primarily due to its ability to process significantly longer input sequences than RoLLaMA2-7B-Instruct and RoMistral-7B-Instruct, respectively. 
Given that a substantial proportion of the epicrises exceed the 4,096-token limit (see Table \ref{tab:dataset_tokens}), we hypothesize that Phi-4-mini-instruct could offer competitive performance by leveraging a broader context, despite being a smaller model, and without explicit pretraining on the Romanian language.

To adapt the model to the clinical QA setting, we fine-tune it using LoRA, updating only $0.0956\%$ of the parameters via token-level cross-entropy. We employ the same LoRA configuration as for RoLLaMA2-7B-Instruct and RoMistral-7B-Instruct. Thanks to the architectural support of Phi-4 for extended input lengths, we evaluate the model under increasingly longer prompts, namely 2,048, 3,072, 4,096, 8,192, 16,384 and 32,768 tokens. This allows us to assess the relationship between available context and model performance, particularly in scenarios where clinical answers are distributed across epicrises of patients with a long medical history. All prompts follow the \texttt{Q+E+A} format.

\subsection*{Biomedical LLM}

LLaMA3-OpenBioLLM-8B is a domain-specialized model designed to address biomedical and clinical tasks. It is based on the LLaMA3 architecture and has undergone extensive pretraining on a diverse collection of biomedical corpora, including high-quality resources such as PubMed abstracts, full-text clinical trials and biomedical textbooks. Through this targeted pretraining, the model achieves a strong understanding of medical terminology, domain-specific reasoning and structured clinical knowledge.

Despite being pretrained primarily in English, LLaMA3-OpenBioLLM-8B can still offer a valuable perspective when evaluated on Romanian clinical question answering. To adapt the model to our Romanian QA task, we train the model via LoRA (with the same configuration as for the other LLMs), updating only 3,407,872 parameters, which corresponds to $0.0424\%$ of the full model size. We use the \texttt{Q+E+A} prompt format and explore two different prompt lengths, 2,048 and 4,096.

%\vspace{-0.1cm}
\subsection*{Baseline Models}
%\vspace{-0.1cm}

To assess the effectiveness of our fine-tuned models and to better isolate the impact of domain and task adaptation, we include four baselines:

The \textbf{random token selector} is a baselines that randomly selects tokens for the input epicrisis. For this baseline, we run three versions with different input lengths (1,024, 2,048, and 4,096 tokens). For each setup, we randomly select the starting token index (between 0 and the maximum input length) and a span length (between 1 and 3 tokens) to serve as the predicted answer. The answer is extracted from the input context (epicrisis), while entirely ignoring the question. This baseline serves as a trivial reference point, highlighting the complexity of the clinical QA task and providing a lower-bound performance estimate.

The \textbf{majority answer model} is another simple baseline is based on answering with the most prevalent answer in the training set. This model consistently answers with ``Nu'' (``No''), regardless of the posed question or the epicrisis content. While still trivial, the majority answer model can be a more competitive baseline than the random token selector.

For each of the open-source LLMs (RoLLaMA2-7B-Instruct, RoMistral-7B-Instruct, Phi-4-mini-instruct, and LLaMA3-OpenBioLLM-8B), we assess performance without any task-specific fine-tuning. We use the \texttt{Q+E+A} prompt format across all models. The \textbf{zero-shot inference} setup is meant to reflect the out-of-the-box capabilities of distinct types of LLMs. 

In addition to the open-source models, we evaluate two state-of-the-art LLMs that are only accessible through commercial APIs, namely GPT-5.2 and Gemini 3 Flash. We refer to these models as \textbf{zero-shot API-based LLMs}. GPT-5.2 is designed to effectively process long contexts, have strong reasoning capabilities, and support multiple languages. Its reasoning ability is enforced via reinforcement learning, while the multilingual support is ensured by the large-scale pretraining on public data. Gemini 3 Flash is an LLM constructed on top of Gemini 3 Pro, inheriting its multimodal and reasoning capabilities. Its architecture is based on a sparse mixture-of-experts (MoE), which is able to activate only a subset of parameters (experts) per input token. This is achieved by learning to dynamically route tokens to a subset of experts. The thinking levels of Gemini 3 Flash are configured to strike a balance between quality, cost and latency. For both models, we use the same prompt format as before, namely \texttt{Q+E+A}, to ensure a fair comparison with the other models. To control evaluation costs, we compare GPT-5.2 and Gemini 3 Flash with the other LLMs on a randomly sampled subset of 100 test instances. The comparative empirical study is intended to provide a reference point for the zero-shot capabilities of current state-of-the-art LLMs on Romanian clinical question answering.

%\vspace{-0.1cm}
\backmatter
%\vspace{-0.1cm}
\section*{Declarations}

\textbf{Acknowledgments} This research is supported by the project ``Romanian Hub for Artificial Intelligence - HRIA'', Smart Growth, Digitization and Financial Instruments Program, 2021-2027, MySMIS no.~351416. This work was also supported by a grant of the Ministry of Research, Innovation and Digitization, CCCDI - UEFISCDI, project number PN-IV-P6-6.3-SOL-2024-0090, within PNCDI IV.

% \vspace{0.5cm}
% \noindent
% \textbf{Ethics approval} The database was approved by the Ethics Committee of Col\c{t}ea Clinical Hospital where the research was conducted, in accordance with Decision No.~19091, dated October 5th, 2021. The database was also approved by the Ethics Committee of ``Dan Furtun\u{a}'' Medical Center, as per Decision No.~1323, dated 11th December, 2025.

% \vspace{0.5cm}
% \noindent
% \textbf{Consent to participate} The use of anonymized data for research purposes is based on the informed consent of patients, as recorded in the medical charts of Col\c{t}ea Clinical Hospital and ``Dan Furtun\u{a}'' Medical Center, and signed individually by each patient at admission. The entire process complies with the requirements of the General Data Protection Regulation (GDPR), the principles of the Declaration of Helsinki, and the standards of good medical and research practices.

\vspace{0.5cm}
\noindent
\textbf{Consent for publication} The authors give their consent for publication.

\vspace{0.5cm}
\noindent
\textbf{Authors' contributions} A.C. Rogoz performed the experiments and wrote the initial draft. R.T. Ionescu designed the study and revised the manuscript. A.V. Anghel and I.L. Antone-Iordache participated in data collection and curation and wrote the initial draft. S. Coniac and A.I. Ionescu participated in data collection and curation, and revised the manuscript.

\vspace{0.5cm}
\noindent
\textbf{Competing interests} The authors have no conflicts of interest to declare that are relevant to the content of this article.

\vspace{0.5cm}
\noindent
\textbf{Availability of data and materials} The dataset has been made publicly available for non-commercial use under the CC BY-NC-SA 4.0 license agreement, at \url{https://github.com/ana-rogoz/MedQARo}.

\vspace{0.5cm}
\noindent
\textbf{Code availability} The code has been made publicly available for non-commercial use under the CC BY-NC-SA 4.0 license agreement, at \url{https://github.com/ana-rogoz/MedQARo}.

\vspace{0.5cm}
\noindent
\textbf{Open access} This article is licensed under a Creative Commons Attribution 4.0 International License, which permits use, sharing, adaptation, distribution and reproduction in any medium or format, as long as you give appropriate credit to the original author(s) and the source, provide a link to the Creative Commons license, and indicate if changes were made. The images or other third party material in this article are included in the article’s Creative Commons license, unless indicated otherwise in a credit line to the material. If material is not included in the article’s Creative Commons license and your intended use is not permitted by statutory regulation or exceeds the permitted use, you will need to obtain permission directly from the copyright holder. To view a copy of this license, visit \url{https://creativecommons.org/licenses/by/4.0/deed.en}.


% \bibliographystyle{bst/sn-mathphys}
\bibliography{references}% common bib file

@inproceedings{craciun-etal-2025-graf,
    title = "{GRAF}: Graph Retrieval Augmented by Facts for {R}omanian Legal Multi-Choice Question Answering",
    author = "Cr\u{a}ciun, Cristian-George  and
      Sm{\u{a}}du, R{\u{a}}zvan-Alexandru  and
      Cercel, Dumitru-Clementin  and
      Cercel, Mihaela-Claudia",
    booktitle = "Findings of the Association for Computational Linguistics: ACL 2025",
    month = jul,
    year = "2025",
    address = "Vienna, Austria",
    publisher = "Association for Computational Linguistics",
    doi = "10.18653/v1/2025.findings-acl.659",
    pages = "12708--12742",
}

@inproceedings{dima-etal-2024-roqllama,
    title = "{R}o{QL}lama: A Lightweight {R}omanian Adapted Language Model",
    author = "Dima, George-Andrei  and
      Avram, Andrei-Marius  and
      Craciun, Cristian-George  and
      Cercel, Dumitru-Clementin",
    booktitle = "Findings of the Association for Computational Linguistics: EMNLP 2024",
    year = "2024",
    address = "Miami, Florida, USA",
    publisher = "Association for Computational Linguistics",
    doi = "10.18653/v1/2024.findings-emnlp.261",
    pages = "4531--4541",
}

@inproceedings{masala2024vorbecstiromanecsterecipetrain,
      title="{``{V}orbe\c{s}ti Rom\^ane\c{s}te?'' A Recipe to Train Powerful Romanian LLMs with English Instructions}", 
      author={Mihai Masala and Denis C. Ilie-Ablachim and Alexandru Dima and Dragos Corlatescu and Miruna Zavelca and Ovio Olaru and Simina Terian-Dan and Andrei Terian-Dan and Marius Leordeanu and Horia Velicu and Marius Popescu and Mihai Dascalu and Traian Rebedea},
    booktitle = "Findings of the Association for Computational Linguistics: EMNLP 2024",
    year = "2024",
    address = "Miami, Florida, USA",
    publisher = "Association for Computational Linguistics",
    doi = "10.18653/v1/2024.findings-emnlp.681",
    pages = "11632--11647",
}

@article{touvron2023llama,
  title="{LLaMA: Open and Efficient Foundation Language Models}",
  author={Touvron, Hugo and Lavril, Thibaut and Izacard, Gautier and Martinet, Xavier and Lachaux, Marie-Anne and Lacroix, Timoth{\'e}e and Rozi{\`e}re, Baptiste and Goyal, Naman and Hambro, Eric and Azhar, Faisal and others},
  journal={arXiv preprint arXiv:2302.13971},
  year={2023},
doi={10.48550/arXiv.2302.13971}
}

@article{kwiatkowski-etal-2019-natural,
    title = "{Natural Questions: A Benchmark for Question Answering Research}",
    author = "Kwiatkowski, Tom  and
      Palomaki, Jennimaria  and
      Redfield, Olivia  and
      Collins, Michael  and
      Parikh, Ankur  and
      Alberti, Chris  and
      Epstein, Danielle  and
      Polosukhin, Illia  and
      Devlin, Jacob  and
      Lee, Kenton  and
      Toutanova, Kristina  and
      Jones, Llion  and
      Kelcey, Matthew  and
      Chang, Ming-Wei  and
      Dai, Andrew M.  and
      Uszkoreit, Jakob  and
      Le, Quoc  and
      Petrov, Slav",
    journal = "Trans Assoc Comput Linguist",
    volume = "7",
    year = "2019",
    address = "Cambridge, MA",
    publisher = "MIT Press",
    doi = "10.1162/tacl_a_00276",
    pages = "452--466",
}

@inproceedings{rajpurkar-etal-2016-squad,
    title = "{SQ}u{AD}: 100,000+ Questions for Machine Comprehension of Text",
    author = "Rajpurkar, Pranav  and
      Zhang, Jian  and
      Lopyrev, Konstantin  and
      Liang, Percy",
    booktitle = "Proceedings of the 2016 Conference on Empirical Methods in Natural Language Processing",
    month = nov,
    year = "2016",
    address = "Austin, Texas",
    publisher = "Association for Computational Linguistics",
    doi = "10.18653/v1/D16-1264",
    pages = "2383--2392"
}

@inproceedings{liro2021,
  title="{LiRo: Benchmark and leaderboard for Romanian language tasks}",
  author={Stefan Daniel Dumitrescu and Petru Rebeja and Beata Lorincz and Mihaela Gaman and Andrei Avram and Mihai Ilie and Andrei Pruteanu and Adriana Stan and Lorena Rosia and Cristina Iacobescu and Luciana Morogan and George Dima and Gabriel Marchidan and Traian Rebedea and Madalina Chitez and Dani Yogatama and Sebastian Ruder and Radu Tudor Ionescu and Razvan Pascanu and Viorica Patraucean},
  booktitle={Proceedings of the 35th Conference on Neural Information Processing Systems},
  year={2021},
  url={https://datasets-benchmarks-proceedings.neurips.cc/paper/2021/hash/5f93f983524def3dca464469d2cf9f3e-Abstract-round1.html},
 publisher = {Curran Associates, Inc.},
}

@inproceedings{nicolae2021roitd,
  title="{RoITD: Romanian IT Question Answering Dataset}",
  author={Nicolae, Drago\c{s} Constantin and Tufi\c{s}, Dan},
  booktitle={Proceedings of the 16th International Conference on Lingustic Resources and Tools for Natural Language Processsing},
  year={2021},
pages={105--117},
url={https://conferences.info.uaic.ro/consilr/prevEditions/Consilr_2021.pdf}
}

@article{jin2020diseasedoespatienthave,
      title={What Disease does this Patient Have? A Large-scale Open Domain Question Answering Dataset from Medical Exams}, 
      author={Jin, Di and Pan, Eileen and Oufattole, Nassim and Weng, Wei-Hung and Fang, Hanyi and Szolovits, Peter},
  journal={Appl Sci},
  volume={11},
  number={14},
  pages={6421},
  year={2021},
  publisher={MDPI},
doi={10.3390/app11146421}
}

@InProceedings{pmlr-v174-pal22a,
  title = 	 "{MedMCQA: A Large-scale Multi-Subject Multi-Choice Dataset for Medical domain Question Answering}",
  author =       {Pal, Ankit and Umapathi, Logesh Kumar and Sankarasubbu, Malaikannan},
  booktitle = 	 {Proceedings of the Conference on Health, Inference, and Learning},
  pages = 	 {248--260},
  year = 	 {2022},
  volume = 	 {174},
  publisher =    {PMLR},
  url = 	 {https://proceedings.mlr.press/v174/pal22a.html},
}

@inproceedings{jin-etal-2019-pubmedqa,
    title = "{P}ub{M}ed{QA}: A Dataset for Biomedical Research Question Answering",
    author = "Jin, Qiao  and
      Dhingra, Bhuwan  and
      Liu, Zhengping  and
      Cohen, William  and
      Lu, Xinghua",
    booktitle = "Proceedings of the 2019 Conference on Empirical Methods in Natural Language Processing",
    year = "2019",
    address = "Hong Kong, China",
    publisher = "Association for Computational Linguistics",
    doi = "10.18653/v1/D19-1259",
    pages = "2567--2577",
}

@inproceedings{pampari-etal-2018-emrqa,
    title = "emr{QA}: A Large Corpus for Question Answering on Electronic Medical Records",
    author = "Pampari, Anusri  and
      Raghavan, Preethi  and
      Liang, Jennifer  and
      Peng, Jian",
    booktitle = "Proceedings of the 2018 Conference on Empirical Methods in Natural Language Processing",
    year = "2018",
    address = "Brussels, Belgium",
    publisher = "Association for Computational Linguistics",
    doi = "10.18653/v1/D18-1258",
    pages = "2357--2368",
}

@article{clark-etal-2020-tydi,
    title = "{T}y{D}i {QA}: A Benchmark for Information-Seeking Question Answering in Typologically Diverse Languages",
    author = "Clark, Jonathan H.  and
      Choi, Eunsol  and
      Collins, Michael  and
      Garrette, Dan  and
      Kwiatkowski, Tom  and
      Nikolaev, Vitaly  and
      Palomaki, Jennimaria",
    journal = "Trans Assoc Comput Linguist",
    volume = "8",
    year = "2020",
    address = "Cambridge, MA",
    publisher = "MIT Press",
    doi = "10.1162/tacl_a_00317",
    pages = "454--470",
}

@inproceedings{lewis-etal-2020-mlqa,
    title = "{MLQA}: Evaluating Cross-lingual Extractive Question Answering",
    author = "Lewis, Patrick  and
      Oguz, Barlas  and
      Rinott, Ruty  and
      Riedel, Sebastian  and
      Schwenk, Holger",
    booktitle = "Proceedings of the 58th Annual Meeting of the Association for Computational Linguistics",
    year = "2020",
    address = "Online",
    publisher = "Association for Computational Linguistics",
    doi = "10.18653/v1/2020.acl-main.653",
    pages = "7315--7330",
}

@inproceedings{artetxe-etal-2020-cross,
    title = "On the Cross-lingual Transferability of Monolingual Representations",
    author = "Artetxe, Mikel  and
      Ruder, Sebastian  and
      Yogatama, Dani",
    booktitle = "Proceedings of the 58th Annual Meeting of the Association for Computational Linguistics",
    month = jul,
    year = "2020",
    address = "Online",
    publisher = "Association for Computational Linguistics",
    doi = "10.18653/v1/2020.acl-main.421",
    pages = "4623--4637",
}

@inproceedings{hu2021lora,
  title="{LoRA: Low-Rank Adaptation of Large Language Models}",
  author={Hu, Edward J. and Shen, Yelong and Wallis, Phillip and Allen-Zhu, Zeyuan and Li, Yuanzhi and Wang, Lu and Chen, Weizhu},
  booktitle = {Proceedings of the International Conference on Learning Representations},
  year={2022},
url={https://openreview.net/forum?id=nZeVKeeFYf9}
}

@article{phi4mini,
  title = "{Phi-4-Mini Technical Report: Compact yet Powerful Multimodal Language Models via Mixture-of-LoRAs}",
 author={Abouelenin, Abdelrahman and Ashfaq, Atabak and Atkinson, Adam and Awadalla, Hany and Bach, Nguyen and Bao, Jianmin and Benhaim, Alon and Cai, Martin and Chaudhary, Vishrav and Chen, Congcong and others},
  journal={arXiv preprint arXiv:2503.01743},
  year={2025},
doi={10.48550/arXiv.2503.01743}
}

@inproceedings{kopf2024openassistant,
  author={K{\"o}pf, Andreas and Kilcher, Yannic and von R{\"u}tte, Dimitri and Anagnostidis, Sotiris and Tam, Zhi Rui and Stevens, Keith and Barhoum, Abdullah and Nguyen, Duc and Stanley, Oliver and Nagyfi, Rich{\'a}rd and others},
title = "{OpenAssistant conversations -- Democratizing large language model alignment}",
year = {2023},
publisher = {Curran Associates Inc.},
booktitle = {Proceedings of the 37th International Conference on Neural Information Processing Systems},
articleno = {2064},
numpages = {13},
location = {New Orleans, LA, USA},
pages={47669--47681},
url={https://proceedings.neurips.cc/paper_files/paper/2023/hash/949f0f8f32267d297c2d4e3ee10a2e7e-Abstract-Datasets_and_Benchmarks.html}
}

@article{mukherjee2023orca,
      title="{Orca: Progressive Learning from Complex Explanation Traces of GPT-4}", 
      author={Subhabrata Mukherjee and Arindam Mitra and Ganesh Jawahar and Sahaj Agarwal and Hamid Palangi and Ahmed Awadallah},
      year={2023},
journal={arXiv preprint arXiv:2306.02707},
doi={10.48550/arXiv.2306.02707}
}

@article{OpenBioLLMs,
  author = {Ankit Pal and Malaikannan Sankarasubbu},
  title = "{OpenBioLLMs: Advancing Open-Source Large Language Models for Healthcare and Life Sciences}",
  year = {2024},
  publisher = {Hugging Face},
  journal = {Hugging Face Repository},
  url = {https://huggingface.co/aaditya/OpenBioLLM-Llama3-70B}
}

@article{alpaca,
  author = {Rohan Taori and Ishaan Gulrajani and Tianyi Zhang and Yann Dubois and Xuechen Li and Carlos Guestrin and Percy Liang and Tatsunori B. Hashimoto},
  title ="{Stanford Alpaca: An Instruction-following LLaMA model}",
  year = {2023},
  publisher = {GitHub},
  journal = {GitHub repository},
  url={https://github.com/tatsu-lab/stanford_alpaca},
}

@inproceedings{papineni2002bleu,
  title="{BLEU: A method for automatic evaluation of machine translation}",
  author={Papineni, Kishore and Roukos, Salim and Ward, Todd and Zhu, Wei-Jing},
  booktitle={Proceedings of the 40th Annual Meeting of the Association for Computational Linguistics},
  pages={311--318},
  year={2002},
  organization={Association for Computational Linguistics},
doi={10.3115/1073083.1073135},
}

@inproceedings{loshchilov2019decoupled,
  title     = "{Decoupled Weight Decay Regularization}",
  author    = {Loshchilov, Ilya and Hutter, Frank},
  booktitle = {Proceedings of the International Conference on Learning Representations},
  year      = {2019},
  url       = {https://openreview.net/forum?id=Bkg6RiCqY7}
}

@article{chowdhery2023palm,
  title="{PaLM: Scaling language modeling with pathways}",
  author={Chowdhery, Aakanksha and Narang, Sharan and Devlin, Jacob and Bosma, Maarten and Mishra, Gaurav and Roberts, Adam and Barham, Paul and Chung, Hyung Won and Sutton, Charles and Gehrmann, Sebastian and others},
  journal={J Mach Learn Res},
  volume={24},
  number={240},
  pages={1--113},
  year={2023},
url={https://jmlr.org/papers/v24/22-1144.html}
}

@inproceedings{hendrycksmeasuring,
  title={Measuring Massive Multitask Language Understanding},
  author={Hendrycks, Dan and Burns, Collin and Basart, Steven and Zou, Andy and Mazeika, Mantas and Song, Dawn and Steinhardt, Jacob},
  booktitle={Proceedings of the International Conference on Learning Representations},
year={2021},
url={https://openreview.net/forum?id=d7KBjmI3GmQ}
}

@article{openai2023gpt4,
  title="{GPT-4 Technical Report}",
  author={Achiam, Josh and Adler, Steven and Agarwal, Sandhini and Ahmad, Lama and Akkaya, Ilge and Aleman, Florencia Leoni and Almeida, Diogo and Altenschmidt, Janko and Altman, Sam and Anadkat, Shyamal and others},
  journal={arXiv preprint arXiv:2303.08774},
  year={2023},
doi={10.48550/arXiv.2303.08774}
}

@article{singhal2023large,
  title={Large language models encode clinical knowledge},
  author={Singhal, Karan and Azizi, Shekoofeh and Tu, Tao and Mahdavi, S Sara and Wei, Jason and Chung, Hyung Won and Scales, Nathan and Tanwani, Ajay and Cole-Lewis, Heather and Pfohl, Stephen and others},
  journal={Nature},
  volume={620},
  number={7972},
  pages={172--180},
  year={2023},
  publisher={Nature Publishing Group},
url={https://www.nature.com/articles/s41586-023-06291-2}
}

@article{singhal2023towards,
  title={Towards expert-level medical question answering with large language models},
  author={Singhal, Karan and Tu, Tao and Gottweis, Juraj and Sayres, Rory and Wulczyn, Ellery and Hou, Le and Clark, Kevin and Pfohl, Stephen and Cole-Lewis, Heather and Neal, Darlene and others},
  journal={Nat Med},
volume={31},
pages={943--950},
  year={2025},
  publisher={Nature Publishing Group},
doi={10.1038/s41591-024-03423-7}
}

@inproceedings{brown2020language,
  title={Language models are few-shot learners},
  author={Brown, Tom and Mann, Benjamin and Ryder, Nick and Subbiah, Melanie and Kaplan, Jared D. and Dhariwal, Prafulla and Neelakantan, Arvind and Shyam, Pranav and Sastry, Girish and Askell, Amanda and others},
booktitle = {Proceedings of the 34th International Conference on Neural Information Processing Systems},
  volume={33},
  pages={1877--1901},
  year={2020},
publisher = {Curran Associates Inc.},
url={https://papers.nips.cc/paper/2020/hash/1457c0d6bfcb4967418bfb8ac142f64a-Abstract.html}
}

@inproceedings{banerjee2005meteor,
  title     = "{METEOR: An Automatic Metric for MT Evaluation with Improved Correlation with Human Judgments}",
  author    = {Banerjee, Satanjeev and Lavie, Alon},
  booktitle = {Proceedings of the ACL Workshop on Intrinsic and Extrinsic Evaluation Measures for MT and/or Summarization},
  pages     = {65--72},
  year      = {2005},
  address   = {Ann Arbor, Michigan},
  publisher = {Association for Computational Linguistics},
url={https://aclanthology.org/W05-0909/}
}

@article{jiang2023mistral7b,
      title="{Mistral 7B}", 
      author={Albert Q. Jiang and Alexandre Sablayrolles and Arthur Mensch and Chris Bamford and Devendra Singh Chaplot and Diego de las Casas and Florian Bressand and Gianna Lengyel and Guillaume Lample and Lucile Saulnier and Lélio Renard Lavaud and Marie-Anne Lachaux and Pierre Stock and Teven Le Scao and Thibaut Lavril and Thomas Wang and Timothée Lacroix and William El Sayed},
      year={2023},
journal={arXiv preprint arXiv:2310.06825},
      doi={10.48550/arXiv.2310.06825}, 
}

@article{Dong-G-2012,
  title="{Prevalence of self-neglect across gender, race, and socioeconomic status: findings from the Chicago Health and Aging Project}",
  author={Dong, XinQi and Simon, Melissa A and Evans, Denis A},
  journal={Gerontology},
  volume={58},
  number={3},
  pages={258--268},
  year={2012},
  publisher={S. Karger AG Basel, Switzerland},
doi={https://doi.org/10.1159/000334256}
}

@article{Bonhomme-JMHG-2007,
author = {Bonhomme, Jean J.},
title = {Men's health: impact on women, children and society},
journal = {J Mens Health Gend},
volume = {4},
number = {2},
pages = {124--130},
year = {2007},
doi = {10.1016/j.jmhg.2007.01.011},
}

@article{ GPT52-2025,
author={OpenAI},
title="{Update to GPT-5 System Card: GPT-5.2}",
  year={2025},
journal={Technical Report},
url = {https://cdn.openai.com/pdf/3a4153c8-c748-4b71-8e31-aecbde944f8d/oai\_5\_2\_system-card.pdf},
accessed={December 2025}
}

@article{ Gemini3-2025,
  title="{Gemini 3 Flash - Model Card}",
  author={Google},
  year={2025},
journal={Technical Report},
url = {https://storage.googleapis.com/deepmind-media/Model-Cards/Gemini-3-Flash-Model-Card.pdf}
}

@article{ Barbero-Arxiv-2025,
  title="{Why do LLMs attend to the first token?}",
  author={Barbero, Federico and Arroyo, Alvaro and Gu, Xiangming and Perivolaropoulos, Christos and Bronstein, Michael and Veli{\v{c}}kovi{\'c}, Petar and Pascanu, Razvan},
  journal={arXiv preprint arXiv:2504.02732},
  year={2025},
doi={10.48550/arXiv.2504.02732}
}

%\clearpage

\end{document}